%% file: main_arxiv.tex
\pdfoutput=1 %
\documentclass{article}
\usepackage{iclr2027_conference,times}
\iclrfinalcopy %
\usepackage[utf8]{inputenc}
\usepackage[T1]{fontenc}
\usepackage{graphicx}
\usepackage{booktabs}
\usepackage{array}
\usepackage{multirow}
\usepackage{amsmath,amssymb}
\usepackage{placeins}
\usepackage{wrapfig}
\usepackage{enumitem}
\usepackage{xcolor}
\usepackage{pgfplots}
\usepgfplotslibrary{groupplots}
\pgfplotsset{compat=1.18}
\definecolor{TaffyCitationBlue}{HTML}{5B9BD5}
\usepackage{hyperref}
\hypersetup{colorlinks=true,citecolor=TaffyCitationBlue,linkcolor=black,urlcolor=black}

\newcommand{\method}{\textsc{Taffy}}
\newcommand{\methodlogo}{\raisebox{-0.22\height}{\includegraphics[height=1.18em]{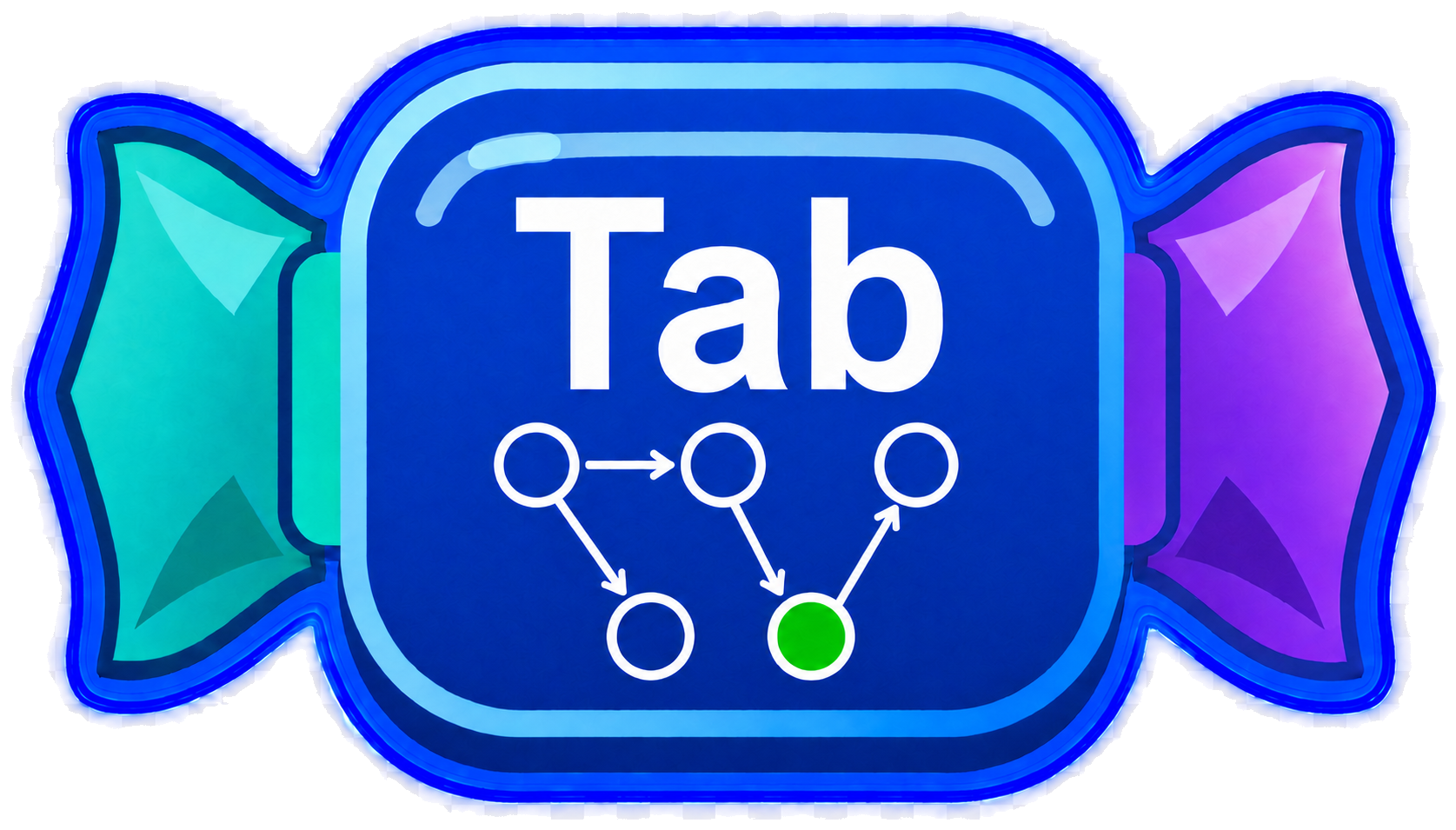}}}

\title{\raggedright\methodlogo\hspace{0.32em}\method: A Task-Adaptive Tabular Foundation Model with In-Context Diversity}

\author{%
Zijian Li\textsuperscript{1,2}, Xiangchen Song\textsuperscript{1}, Gongxu Luo\textsuperscript{2}, Jie Qiao\textsuperscript{3}, Ruichu Cai\textsuperscript{3}, \\
\textbf{Zhenhao Chen\textsuperscript{2}, Xinshuai Dong\textsuperscript{1}, Fan Feng\textsuperscript{4}, Guangyi Chen\textsuperscript{1,2}, Kun Zhang\textsuperscript{1,2}} \\
\textsuperscript{1}Carnegie Mellon University \quad
\textsuperscript{2}Mohamed bin Zayed University of Artificial Intelligence \\
\textsuperscript{3}Guangdong University of Technology \quad
\textsuperscript{4}University of California, San Diego
}

\begin{document}\maketitle
\lhead{Preprint}

\begin{abstract}
Recent progress in tabular foundation models suggests that training on synthetic tasks can substantially improve in-context learning capabilities, with overall performance largely depending on how well models can infer task-specific predictive relationships from the available context during inference. In this paper, we introduce \textbf{\method}, a tabular foundation model with an In-Context Diversity Prior and a Task-Conditioned Looped Transformer that strengthen this ability. Specifically, to construct each synthetic pretraining context, the In-Context Diversity Prior samples from multiple related environments derived via controlled interventions and distribution shifts on a shared causal process. This in-context diversity encourages the model to learn a more comprehensive and task-specific representation. Moreover, the Task-Conditioned Looped Transformer iteratively and selectively applies a shared group of Transformer blocks to refine contextual representations, with a task-conditioned gate modulating the final hidden-state update. This enables task-adaptive iterative refinement. Together, these components encourage the model to identify predictive relationships from contextual contrasts during pretraining and dynamically modulate context integration for each task. Across six classification and five regression benchmark datasets, \textbf{\method} attains the lowest average rank.
\end{abstract}

\input{Taffy_intro_section}

\input{Taffy_related_work}

\input{Taffy_problem_formulation}

\input{Taffy_method}

\input{Taffy_experiments}

\input{Taffy_conclusion}

\bibliography{references,additional_references}
\bibliographystyle{iclr2027_conference}

\clearpage
\appendix
\input{Taffy_appendix}

\end{document}

%% file: Taffy_intro_section.tex
\section{Introduction}
\label{sec:introduction}

Tabular foundation models leverage in-context learning (ICL) to predict query labels from labeled support examples without explicit parameter updates. Prior-data fitted networks formulate this paradigm as approximating Bayesian inference under a task prior: synthetic tasks provide support--query pairs, and pretraining optimizes query predictions conditioned on the support set \citep{muller2022transformers}.
Recent architectures such as TabPFN, TabICL, LimiX, and Mitra have advanced tabular ICL through developments in synthetic data generation and prediction architectures \citep{hollmann2023tabpfn,hollmann2025accurate,qu2025tabicl,qu2026tabiclv2,zhang2025limix,wang2026limix2m,zhang2025mitra}.
Because semantic meanings and feature dependencies shift across datasets, a model cannot rely on fixed feature semantics; instead, it must infer task-specific functional dependencies directly from the support context---an intuition formalized by recent kernel and nearest-neighbor views of tabular ICL \citep{koshil2025context,miftachov2026interpretable}.
Consequently, generalization to unseen tasks depends on how effectively the model learns task-specific predictive relationships from context.

To strengthen this ability, we introduce \textbf{\method}, which combines an \emph{In-Context Diversity Prior} for synthetic task construction with a \emph{Task-Conditioned Looped Transformer} for task-dependent context processing. First, the \emph{In-Context Diversity Prior} incorporates observations from multiple related environments into a single synthetic pretraining context. Starting from a randomly sampled causal prototype, we first generate a set of environments through controllable distribution shifts \citep{zhang2015multisource} and interventions \citep{pearl2009causality}, then combine their support data and query data to form one prediction task. This design is inspired by causal discovery under changing distributions \citep{huang2020heterogeneous} and latent-variable learning with auxiliary information \citep{hyvarinen2019nonlinear}, which allow the model to explore a sufficient amount of variation across different data generation conditions.
Although the model is not provided with environment identifiers or access to the generating process, the mixing prior enables observations from related conditions which provides contrasting evidence about predictive relationships.
Training on these contrasts encourages \method{} to infer task-specific predictive relationships from the available context.

\begin{figure}[t]
    \centering
    \includegraphics[width=\linewidth]{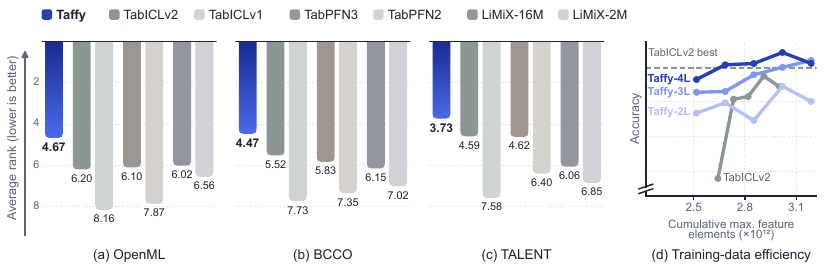}
    \vspace{-5mm}
    \caption{Predictive performance and pretraining-data efficiency. (a--c) Average accuracy ranks on OpenML-CC18, BCCO, and TALENT using the ranking pools of Table~\ref{tab:main-performance} (lower is better; rank axes start at 2). (d) Accuracies on TALENT classification datasets versus cumulative maximum feature elements. Markers denote measured checkpoints connected by lines; the grey line marks TabICLv2's best measured accuracy.}
    
    \label{fig:main-results}
    \label{fig:benchmark-ranks}
    \vspace{-1mm}
\end{figure}

Second, the \emph{Task-Conditioned Looped Transformer} combines a looped transformer stack with a task-specific gate to strengthen the model's ability to learn task-specific predictive relationships from context. Building on representations produced by the column and row encoders, our model repeatedly applies a shared ICL Transformer stack \citep{dehghani2019universal,yang2024looped} to further integrate information from the labeled context. Later passes recompute attention using representations already informed by support features and labels, allowing them to build on information integrated earlier. Since the usefulness of further refinement may vary across tasks, we introduce a task-conditioning module that maps support-dataset statistics to a task-dependent scalar gate modulating the final hidden-state update. The shared loop enables iterative context processing, while the gate adjusts how strongly the final refinement contributes to prediction with the number of passes held fixed. Since these feature and label statistics are computed solely from the labeled support and unlabeled query dataset, this task-dependent refinement is also applicable when inferring predictive relationships for unseen real-world tasks. Together, the in-context diversity prior provides relational evidence, while the task-conditioned looped Transformer refines the tabular representations.

We evaluate \method{} across six classification and five regression benchmark suites. Using Taffy-4L throughout the classification evaluation, \method{} obtains the lowest average rank on all six classification suites; the regression model ranks first on all five regression suites. Figure~\ref{fig:main-results}(a--c) shows selected foundation models on OpenML-CC18, BCCO, and TALENT, with the full comparison in Table~\ref{tab:main-performance}. Figure~\ref{fig:main-results}(d) compares \method{} with reproduced TabICLv2 \citep{qu2026tabiclv2} across measured pretraining budgets, providing an exploratory view of pretraining data efficiency. Section~\ref{sec:data-efficiency} defines the cumulative maximum-feature-element budget and reports the measured checkpoints.

\noindent\textbf{Our contributions are summarized as follows:}
\begin{itemize}[leftmargin=*,nosep]
    \item We introduce a synthetic prior that places causally related environments within the same task.
    \item We develop a looped Transformer whose final refinement is modulated by task-specific statistics.
    \item \method{} obtains the lowest average rank on all classification and regression benchmarks.
\end{itemize}

%% file: Taffy_related_work.tex
\section{Related Work}
\label{sec:related-work}

\subsection{Tabular Foundation Models}
Tabular foundation models learn from synthetic tasks to make predictions on real-world datasets through in-context learning \citep{qu2025tabicl,zhang2025mitra}. Building on prior-data fitted networks \citep{muller2022transformers}, the TabPFN family extends this approach across classification, regression, and larger tables \citep{hollmann2023tabpfn,hollmann2025accurate,grinsztajn2025tabpfn25}. Generator design includes tree-based tasks, mixtures of priors, and distributional perturbations \citep{denbreejen2024tabforestpfn,zhang2025mitra,bouadi2026shaping}, while the TabICL and LimiX families develop table representations and contextual interactions \citep{qu2025tabicl,qu2026tabiclv2,zhang2025limix,wang2026limix2m}. Other approaches use real-data pretraining \citep{ma2025tabdpt,garg2025realtabpfn} or self-supervised representation learning \citep{wu2024switchtab}. Closely related to our task construction, Drift-Resilient TabPFN generates evolving SCM mechanisms and uses ordered domain indices for temporal extrapolation \citep{helli2024drift}. \method{} instead constructs unordered mixtures of related environments through distribution shifts and interventions, without supplying environment identifiers. Its focus is on placing observations from these conditions within the same support--query task, making their joint use part of the prediction problem practiced during pretraining.

\subsection{Looped Transformers}
Universal Transformers and ALBERT share parameters across processing depth \citep{dehghani2019universal,lan2020albert}, while Relaxed Recursive Transformers add depth-specific low-rank adapters \citep{bae2025relaxed}. In ICL, looped Transformers have been studied as models of iterative algorithms \citep{yang2024looped}, including connections to multi-step preconditioned gradient descent for linear regression \citep{gatmiry2024looped}. Work on latent reasoning explores repeated computation in hidden states \citep{saunshi2025reasoning,geiping2025scaling}, and CoTFormer permits attention to intermediate representations from earlier iterations \citep{mohtashami2025cotformer}. For tabular prediction, \citet{balef2026one} investigate layerwise inference dynamics and a parameter-efficient single-layer looped model. A separate line of work learns halting or token-level computation allocation \citep{graves2016adaptive,banino2021pondernet,raposo2024mixture,chen2025inner,bae2026mixture}. In this paper, \method{} uses shared multi-block recurrence to refine representations already conditioned on each task. A gate based on task statistics modulates only the final update, making its contribution task-dependent without token routing or changing the number of passes.

\subsection{Causal Learning}
Our task prior is related to causal learning through its use of structured changes across environments \citep{scholkopf2021toward}. Causal discovery studies changing mechanisms \citep{huang2020heterogeneous} and cross-environment invariance \citep{peters2016causal}, while multi-source domain adaptation relates distributional variation to transferable mechanisms \citep{zhang2015multisource}. Nonlinear ICA and iVAE establish latent identifiability using auxiliary information under specific assumptions \citep{hyvarinen2019nonlinear,khemakhem2020ivae}. Related work considers latent causal identification from interventions \citep{varici2024general} and content identification from paired augmentations \citep{vonkugelgen2021self}. These results motivate studying structured variation, but do not establish identifiability for our mixtures without environment identifiers. In \method{}, causal prototypes and interventions serve as tools for constructing related environments within the same pretraining task. The goal is to train prediction from their pooled observations, rather than to recover the generating process.

%% file: Taffy_problem_formulation.tex
\section{Problem Formulation}
\label{sec:problem-formulation}
\label{sec:problem}

\textbf{Tabular Foundation Model with in-context learning.}
We consider classification and regression tasks on tabular data.
For a given task, each sample has a feature vector
$x \in \mathbb{R}^{d}$ and the corresponding label $y \in \mathcal{Y}$, where
$\mathcal{Y}=\{1,\ldots,K\}$ for classification and
$\mathcal{Y}=\mathbb{R}$ for regression. The available data consist of a labeled support dataset
$S=\{(x_i^s,y_i^s)\}_{i=1}^{n_s}$ and unlabeled query dataset $X_Q=(x_j^q)_{j=1}^{n_q}$ whose corresponding labels
$y_Q=(y_j^q)_{j=1}^{n_q}$ are unknown to the model.  A tabular foundation model with parameters $\theta$ predicts
the query labels by conditioning on the support dataset:
\begin{equation}
\small
    (\hat{y}_1,\ldots,\hat{y}_{n_q})=f_{\theta}(X_Q;S),
    \label{eq:icl-prediction}
\end{equation}
where $\hat{y}_j$ denotes the predictive output for query $x_j^q$.
The output is a class-probability vector for classification
or a vector of conditional quantile predictions for regression.

\textbf{Pretraining objective.}
Following the prior-data fitted framework
\citep{muller2022transformers,qu2026tabiclv2},
we learn $\theta$ from synthetic prediction tasks.
Let $\mathcal{E}=(S,X_Q,y_Q)$ denote a synthetic pretraining
task, and let $\Pi$ denote the distribution over such tasks
induced by the synthetic prior.
Pretraining minimizes the expected query prediction loss as follows:
\begin{equation}
\small
    \mathcal{L}_{\mathrm{pre}}(\theta)
    =
    \mathbb{E}_{\mathcal{E}\sim\Pi}
    \left[
        \frac{1}{n_q}
        \sum_{j=1}^{n_q}
        \ell_{\mathcal{E}}
        \left(
            [f_{\theta}(X_Q;S)]_j,
            y_j^q
        \right)
    \right],
    \label{eq:pretraining-objective}
\end{equation}
where $\ell_{\mathcal{E}}$ is defined separately for classification
and regression, following TabICLv2 \citep{qu2026tabiclv2}.
For classification, we use cross-entropy loss.
For regression, we use pinball loss averaged over 999 predicted
quantiles. At inference, point predictions are obtained by
averaging these quantiles. Query labels supervise the pretraining objective but are excluded from the model inputs. The prior $\Pi$ specifies the joint construction of support and query data.

%% file: Taffy_method.tex
\section{Taffy}
\label{sec:method}

\subsection{Overall Framework}
\label{sec:overview}

\begin{figure}[!t]
  \centering
  \includegraphics[width=\linewidth]{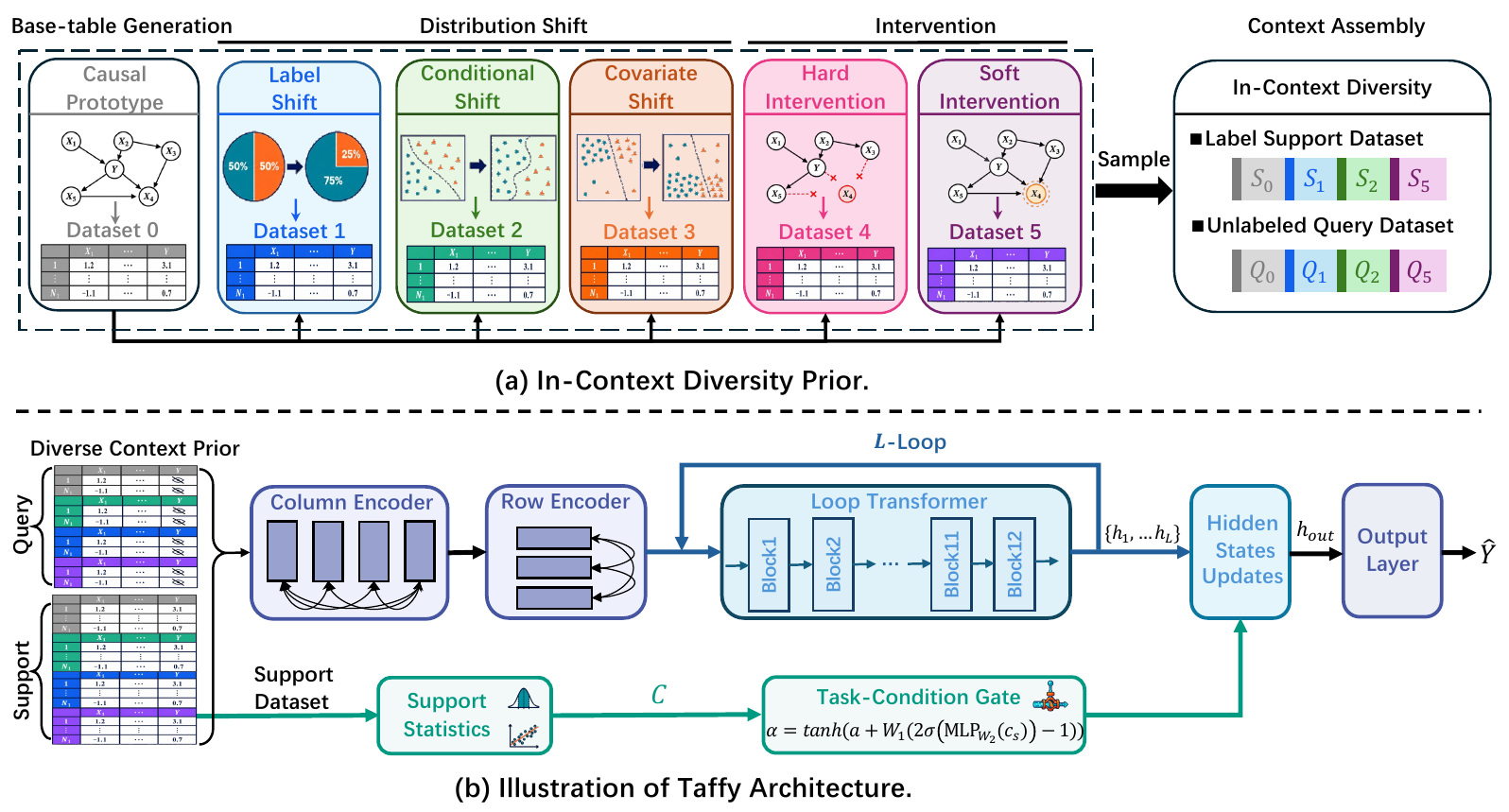}
  \caption{The pretraining pipeline of \method{} combines the in-context diversity prior with the task-conditioned looped transformer.
(a) In-Context Diversity Prior proceeds through three steps: base-table generation, multiple-environment construction, and context assembly.
(b) Column and row encoders produce representations for a looped Transformer, whose final update is modulated by a support-derived task gate before query prediction.}
  \label{fig:taffy-overview}
\end{figure}

\method{} is designed to learn task-specific predictive relationships from a new support context. Figure~\ref{fig:taffy-overview} shows how the training-task construction and prediction network serve this goal. Specifically, the \emph{In-Context Diversity Prior} defines the contextual prediction problems used during pretraining, including tasks that combine observations from multiple causally related environments. The \emph{Task-Conditioned Looped Transformer} refines support-conditioned representations through a reusable update of Transformers, with later passes building on information integrated by earlier ones. The query-label objective in Eq.~\eqref{eq:pretraining-objective} trains the network to predict from each sampled task's support data.

Given a labeled support dataset and query features, column and row encoders produce row representations that incorporate support-label information. The looped Transformer applies a shared ICL stack repeatedly to these representations. A separate branch summarizes support features and labels statistically and computes a scalar gate, which modulates the final recurrent update before query prediction. At downstream inference, only the prediction network and its task-conditioning branch are used; the synthetic generator and environment identifiers of synthetic data are not required.

\subsection{In-Context Diversity Prior}
\label{sec:prior}

\paragraph{Design principle.}
The synthetic prior guides how the model learns feature--label predictive relationships from context. Our \emph{In-Context Diversity Prior} places data from multiple causally related environments within the same task. Starting from a shared causal prototype, we construct these environments through different types of distribution shifts and interventions. Because the environments share the same causal prototype but slightly different mechanisms, their joint context provides contrasting evidence about how feature--label relationships behave under changing conditions. In causal discovery, such mechanism diversity can help identify causal structures among variables that are difficult to identify from a single observational distribution \citep{zhang2015multisource,pearl2009causality,huang2020heterogeneous}. We adapt this insight to synthetic pretraining by using multi-environment diversity to help the model learn predictive relationships from context, rather than to recover the underlying causal graph. Figure~\ref{fig:taffy-overview}(a) summarizes the construction in three steps: base-table generation, multiple-environment construction, and context assembly, which are illustrated as follows.

\textbf{Step 1: Base-table generation.}
We first sample a causal process \citep{hollmann2023tabpfn} as the causal prototype that defines a joint distribution over features and labels, then draw a base table from this distribution, serving as the starting point for constructing the causally related environments.

\textbf{Step 2: Multiple-environment construction.}
Starting from the shared causal prototype, we construct multiple environments through distribution shifts in data sampling and interventions on structural mechanisms. Let $D$ denote the number of environments, with $(S_e,Q_e)$ denoting the support and query datasets for environment $e$. We construct datasets for different environments as follows.

\emph{Distribution shifts.}
Keeping the sampled causal graph and its structural mechanisms fixed, we first generate a fresh candidate table for each environment. We then construct environment-specific datasets through three sampling schemes \citep{zhang2015multisource} corresponding to three types of distribution shifts. For \emph{label shift}, we change $p(y)$ relative to the base table while keeping $p(x\mid y)$ unchanged. We do so by assigning different sampling probabilities to the labels and sampling data points based on their labels. For \emph{covariate shift}, we change $p(x)$ relative to the base table while keeping $p(y\mid x)$ unchanged. We achieve it by assigning sampling probabilities based on feature values and sampling data points accordingly. For \emph{conditional shift}, we change $p(x\mid y)$ relative to the base table while keeping $p(y)$ unchanged. Specifically, we implement it by assigning feature-dependent sampling probabilities separately within each label and sampling data points accordingly. Definitions of these shifts and implementation details are provided in Appendix~\ref{app:distribution-shifts}.

\emph{Interventions.}
Starting from the same sampled causal graph in step 1, we construct interventional environments by modifying the mechanisms of selected non-label variables \citep{pearl2009causality}, without directly intervening on the label node. For a \emph{hard intervention}, we replace each selected variable's structural equation with an externally specified value that does not depend on its parents. This removes its incoming edges, while its outgoing effects remain. For a \emph{soft intervention}, we keep the causal graph fixed but replace the selected variable's structural mechanism with an environment-specific, potentially nonlinear mechanism over the same parents, thereby changing how the parents determine the variable without removing their influence \citep{brouillard2020differentiable}. We then generate a new table from the intervened causal process, propagating intervention effects to descendant variables. Definitions and implementation details are provided in Appendix~\ref{app:interventions}.

\textbf{Step 3: Context assembly.}
We combine environment-specific support and query data as follows:
\begin{equation}
\small
    S=\biguplus_{e=1}^{D}S_e,
    \qquad
    Q=\biguplus_{e=1}^{D}Q_e,
    \label{eq:context-assembly}
\end{equation}
where $\biguplus$ denotes row-wise concatenation with repeated examples retained. The resulting $S$ and $Q=(X_Q,y_Q)$ form the assembled support and query datasets. For multi-environment tasks, we first sample $D$ (up to 4 distinct environments, including the unchanged base-table environment), then sample the changes used to construct the remaining environments from the candidate distribution shifts and interventions in Step 2. The model receives the labeled support dataset and query features, while query labels provide supervision through the pretraining objective. To retain exposure to single-environment tasks that may arise during downstream inference, we leave $50\%$ of pretraining tasks as single-environment tasks and apply the multi-environment construction to the remainder.

It is noted that the environment indices are used only to assemble synthetic tasks and are not supplied to the model. The predictor learns from the pooled support and query data through query-label supervision, with no environment-prediction or causal-identification objective.

\subsection{Column and Row Encoders}
\label{sec:encoders}

We follow the column-then-row encoding design of TabICLv2 \citep{qu2026tabiclv2} to obtain fixed-dimensional representations from tables with varying feature counts. Circular shifts form overlapping groups of three feature values, each projected to a 128-dimensional token by a shared linear layer. Support-label embeddings are added to the corresponding support tokens to incorporate feature--label associations. Three induced-attention blocks then use 128 inducing vectors to summarize support information for each feature group and propagate it to support and query tokens.

Next, the row encoder models interactions among the feature-group tokens within each example using three Transformer blocks and four learned \texttt{[CLS]} tokens. Concatenating the four \texttt{[CLS]} outputs yields a 512-dimensional row representation. A separate label embedding is then added to each support-row representation before the looped ICL learner. These augmented support representations and the query-row representations form the initial hidden states $h^{(0)}$. Query labels are never used as inputs. Both encoders run once per forward pass; only the ICL learner is repeated.

\subsection{Task-Conditioned Looped Transformer}
\label{sec:loop}

\textbf{Design principle.}
The In-Context Diversity Prior exposes the model to varied predictive relationships by placing observations from multiple environments within the same context. These relationships differ in complexity: some can be identified from the initial context representations, whereas others benefit from further integration across examples and features. Looped computation with task-conditioned gating provides a way to accommodate this variation. Reapplying a shared Transformer allows each pass to build on contextual information integrated by earlier passes \citep{dehghani2019universal,yang2024looped}, while a task-specific gate regulates how much the final refinement contributes to prediction. The module therefore adapts the influence of iterative computation to the current task, while keeping the number of passes fixed and sharing parameters across passes.

\textbf{Looped Transformer.}
After column and row encoding in Section \ref{sec:encoders}, we obtain the initial representations $h^{(0)}$. We then use a looped ICL Transformer to progressively integrate information across the labeled context. Let $F_\theta$ denote the twelve-block ICL Transformer with hidden width 512. Following shared recurrent computation of Transformers~\citep{dehghani2019universal,yang2024looped,balef2026one}, we apply the same stack to successive hidden states:
\begin{equation}
\small
    h^{(\ell)}=F_\theta(h^{(\ell-1)};M_S),\qquad \ell=1,\ldots,L,
    \label{eq:shared-loop}
\end{equation}
where $L$ is the number of passes and $M_S$ denotes a support-only attention pattern. In each block, all $N=n_s+n_q$ positions serve as queries, whereas only the first $n_s$ support positions serve as keys and values. We implement this pattern by placing support representations before query representations and slicing the first $n_s$ hidden states for the key/value sequence. Thus, both support and query representations are updated, while query examples do not attend to one another. 

\textbf{Task-specific gate.}
The task-specific gate makes the final recurrent update adaptively depend on a statistical description of the current task. Specifically, we summarize the support dataset using statistics such as feature means, standard deviations, and class proportions, and combine them with input-shape information into a parameter-free descriptor $c_S\in\mathbb{R}^{51}$ to guide contextual refinement. The complete list of statistics is provided in Appendix~\ref{app:support-statistics}. Since query labels are unavailable in real-world tasks, we compute the label summaries solely from the labeled support dataset. For regression, the four class-specific entries are set to zero.
To translate this task descriptor into a gate that regulates the final refinement, we use a scalar-output multilayer perceptron (MLP) parameterized by $W_2$ to compute a bounded task-dependent correction to a global scalar $a$:
\begin{equation}
\small    b_S=W_1\bigl(2\sigma(\operatorname{MLP}_{W_2}(c_S))-1\bigr),
    \qquad \alpha_S=\tanh(a+b_S),
    \label{eq:support-gate}
\end{equation}
where $\sigma$ is the logistic sigmoid, $W_1$ is a scalar scaling coefficient, and $W_2$ denotes the parameters of the MLP. The gate combines a global update scale with a task-specific correction.

\textbf{Conditioned final update.}
After $L$ passes, the looped Transformer produces hidden states $h^{(1)},\ldots,h^{(L)}$, each containing the support and query representations. We use the task-specific gate to scale the refinement from the final pass and add it to the preceding state:
\begin{equation}
\small
    h^{\mathrm{out}}=h^{(L-1)}+
    \alpha_S\bigl(h^{(L)}-h^{(L-1)}\bigr).
    \label{eq:conditioned-update}
\end{equation}
Here, $h^{(L)}-h^{(L-1)}$ is the update contributed by the final pass. Setting $\alpha_S=0$ retains the preceding state; a positive gate preserves a fraction of this update, while a negative gate reverses its direction. All $L$ passes are executed, so task conditioning changes the final representation rather than the number of computation steps. Finally, the prediction module produces $\widehat{Y}_Q=g_\theta(h^{\mathrm{out}}_Q)$, where $g_\theta$ maps query representations to class probabilities or conditional quantile predictions for regression.

%% file: Taffy_experiments.tex
\section{Experiments}
\label{sec:experiments}

\begin{table}[!t]
\vspace{-5mm}
\centering
\caption{Average ranks across all evaluated classification and regression datasets (lower is better). The classification row uses Taffy-4L on every suite; regression uses the separately trained model.}
\label{tab:main-performance}
\begingroup
\fontsize{8}{9}\selectfont
\setlength{\tabcolsep}{1.5pt}
\renewcommand{\arraystretch}{0.90}
\setlength{\aboverulesep}{1.1pt}
\setlength{\belowrulesep}{1.1pt}
\begin{tabular*}{\linewidth}{@{\extracolsep{\fill}}c|*{6}{c}|*{5}{c}@{}}
\toprule
\multirow{2}{*}[\dimexpr-\aboverulesep/2-\cmidrulewidth/2-\belowrulesep/2\relax]{\textbf{Model}} & \multicolumn{6}{c|}{\textbf{Classification}} & \multicolumn{5}{c}{\textbf{Regression}} \\
\cmidrule(lr){2-7}\cmidrule(l){8-12}
& {\fontsize{7}{8}\selectfont\textbf{BCCO}}
& {\fontsize{7}{8}\selectfont\textbf{OpenML}}
& {\fontsize{7}{8}\selectfont\textbf{PFN}}
& {\fontsize{7}{8}\selectfont\textbf{TALENT}}
& {\fontsize{7}{8}\selectfont\textbf{TabArena}}
& {\fontsize{7}{8}\selectfont\textbf{TabZilla}}
& {\fontsize{7}{8}\selectfont\textbf{BCCO}}
& {\fontsize{7}{8}\selectfont\textbf{CTR23}}
& {\fontsize{7}{8}\selectfont\textbf{PFN}}
& {\fontsize{7}{8}\selectfont\textbf{TALENT}}
& {\fontsize{7}{8}\selectfont\textbf{TabArena}} \\
\midrule
\multicolumn{12}{c}{\textbf{Ensemble methods}} \\
\midrule
AutoGluon & 10.62 & 8.29 & 9.88 & 9.96 & 9.98 & 8.52 & 5.96 & 5.85 & 5.96 & 5.89 & 6.15 \\
\midrule
\multicolumn{12}{c}{\textbf{Tree-based methods}} \\
\midrule
CatBoost & 10.29 & 9.88 & 9.30 & 10.70 & 10.78 & 11.02 & 7.20 & 7.39 & 7.25 & 7.02 & 7.77 \\
XGBoost & 11.66 & 10.52 & 9.95 & 11.81 & 11.03 & 11.07 & 11.30 & 10.55 & 10.93 & 10.39 & 10.92 \\
ET & 12.23 & 11.55 & 10.96 & 13.34 & 14.77 & 11.83 & 10.15 & 10.03 & 10.29 & 9.76 & 9.54 \\
RF & 10.66 & 13.45 & 11.38 & 13.06 & 10.48 & 13.15 & 10.61 & 11.48 & 10.32 & 10.06 & 10.54 \\
\midrule
\multicolumn{12}{c}{\textbf{Deep tabular methods}} \\
\midrule
SwitchTab & 16.73 & 16.46 & 16.16 & 17.45 & 18.98 & 16.00 & 16.85 & 16.36 & 16.39 & 16.47 & 17.23 \\
T2G-Former & 16.43 & 17.70 & 19.00 & 16.85 & 14.95 & 17.06 & 13.24 & 13.88 & 14.43 & 13.42 & 14.23 \\
TANGOS & 14.25 & 16.16 & 16.93 & 15.78 & 14.39 & 14.93 & 12.04 & 11.18 & 12.43 & 12.74 & 13.00 \\
TabCaps & 14.73 & 14.02 & 14.45 & 13.90 & 14.86 & 13.48 & --- & --- & --- & --- & --- \\
TabM & 12.20 & 11.68 & 12.30 & 11.56 & 13.00 & 11.37 & 16.11 & 15.52 & 15.18 & 15.55 & 17.08 \\
TabNet & 18.90 & 18.34 & 19.34 & 18.63 & 18.88 & 18.93 & 16.15 & 15.58 & 16.96 & 16.35 & 16.38 \\
TabR & 11.91 & 11.11 & 11.14 & 11.12 & 12.12 & 10.54 & 10.74 & 10.67 & 12.93 & 11.14 & 12.31 \\
TabTransformer & 16.60 & 13.67 & 13.34 & 15.66 & 16.14 & 14.09 & 18.52 & 17.55 & 16.57 & 17.47 & 17.92 \\
\midrule
\multicolumn{12}{c}{\textbf{Tabular foundation models}} \\
\midrule
TabICLv1 & 7.73 & 8.16 & 7.93 & 7.58 & 7.06 & 8.65 & --- & --- & --- & --- & --- \\
LimiX-2M & 7.02 & 6.56 & 6.14 & 6.85 & 5.72 & 6.20 & 9.76 & 8.88 & 7.75 & 9.17 & 7.15 \\
TabPFN2 & 7.35 & 7.87 & 6.68 & 6.40 & 6.44 & 7.33 & 5.09 & 6.30 & 5.50 & 6.16 & 6.15 \\
Mitra & 9.72 & 12.60 & 12.20 & 11.34 & 10.31 & 12.09 & 9.30 & 10.39 & 8.93 & 10.35 & 9.92 \\
LimiX-16M & 6.15 & 6.02 & 7.00 & 6.06 & 6.03 & 6.67 & 8.09 & 7.55 & 7.57 & 8.00 & 6.46 \\
TabICLv2 & 5.52 & 6.20 & 5.57 & 4.59 & 5.88 & 6.41 & 3.72 & 3.73 & 3.39 & 4.15 & 3.23 \\
TabPFN3 & 5.83 & 6.10 & 6.46 & 4.62 & 4.73 & 6.48 & 3.11 & 4.18 & 3.93 & 3.48 & 2.69 \\
\textbf{\method{}} & \textbf{4.47} & \textbf{4.67} & \textbf{4.89} & \textbf{3.73} & \textbf{4.45} & \textbf{5.19} & \textbf{2.07} & \textbf{2.94} & \textbf{3.29} & \textbf{2.45} & \textbf{1.31} \\
\bottomrule
\end{tabular*}
\endgroup
\vspace{-7pt}
\end{table}

\vspace{-4pt}
\subsection{Experimental Setup}
\label{sec:setup}

\begingroup
\setlength{\parskip}{3pt}
\textbf{Benchmarks.}
We evaluate classification and regression on TALENT \citep{liu2024talenttabularanalyticslearning}, BCCO \citep{zhang2025limix}, the TabPFN benchmark \citep{hollmann2025accurate}, and TabArena \citep{erickson2025tabarena}, plus OpenML-CC18 \citep{bischl2021openml} and TabZilla \citep{mcelfresh2023tabzilla} for classification and CTR23 \citep{fischer2023ctr23} for regression. Preprocessing is detailed in Appendix~\ref{app:experimental-details}.\looseness=-1

\textbf{Baselines and inference.}
Our baseline roster covers four groups: (i) tree-based methods, including CatBoost \citep{prokhorenkova2018catboost}, XGBoost \citep{chen2016xgboost}, random forests (RF) \citep{breiman2001random}, and extremely randomized trees (ET) \citep{geurts2006extremely}; (ii) deep tabular and related predictors, including TabM \citep{gorishniy2025tabm}, TabR \citep{gorishniy2024tabr}, and SwitchTab \citep{wu2024switchtab}; (iii) tabular foundation models, including TabPFN2 \citep{hollmann2025accurate}, TabPFN3 \citep{grinsztajn2026tabpfn3}, TabICLv1/v2 \citep{qu2025tabicl,qu2026tabiclv2}, LimiX-2M/16M \citep{wang2026limix2m,zhang2025limix}, and Mitra \citep{zhang2025mitra}; and (iv) AutoGluon \citep{erickson2020autogluon}. Please refer to Appendix~\ref{app:baseline-roster} for more details of the baselines.

\textbf{Implementation and evaluation.}
We conduct pretraining on a cluster with 192 AMD MI210 GPUs. We train for 25,000 steps with the Muon optimizer. Each pretraining context contains up to four environments, with equal numbers of rows per environment subject to a fixed total row budget. We denote by Taffy-$X$L the model whose shared ICL stack is applied $X$ times. For the primary comparison, classification uses Taffy-4L on all suites; regression uses a separately trained model. Pretraining Taffy-2L, Taffy-3L, and Taffy-4L takes approximately 2.54, 3.07, and 3.65 days, respectively, on 64 GPUs (Appendix~\ref{app:training-cost}). Our primary metric is average rank within each benchmark: methods are ranked on every dataset using classification accuracy or regression RMSE, and the resulting dataset-level ranks are averaged. We further consider the Elo metric \citep{elo1978rating} in Appendix~\ref{app:elo-comparison}.\looseness=-1
\par
\endgroup

\vspace{-4pt}
\subsection{Overall Performance}
\label{sec:performance}
\label{sec:ablations}

\textbf{Main Results.}
Following TabICLv2 \citep{qu2026tabiclv2}, we exclude datasets overlapping with its development set from the primary evaluation and evaluate the 12 TALENT tasks with more than ten classes separately. For these many-class tasks, we use TabICL's hierarchical classification \citep{qu2025tabicl} with mixed-radix ensembling; their rank and Elo results are reported in Appendix~\ref{app:many-class-evaluation}. We further report rank and Elo under the LimiX-2M dataset-selection criteria \citep{wang2026limix2m} in Appendix~\ref{app:limix-compatible-evaluation}. Table~\ref{tab:main-performance} summarizes average ranks across six classification and five regression benchmarks. \method{} obtains the lowest average rank on all six classification and five regression benchmarks. On TALENT, it obtains ranks of 3.73 for classification and 2.45 for regression, compared with 4.62 and 3.48 for TabPFN3 and 4.59 and 4.15 for TabICLv2. 

\begin{figure}[!t]
\vspace{-2mm}
\centering
\scalebox{1}[0.9]{\includegraphics[width=\textwidth]{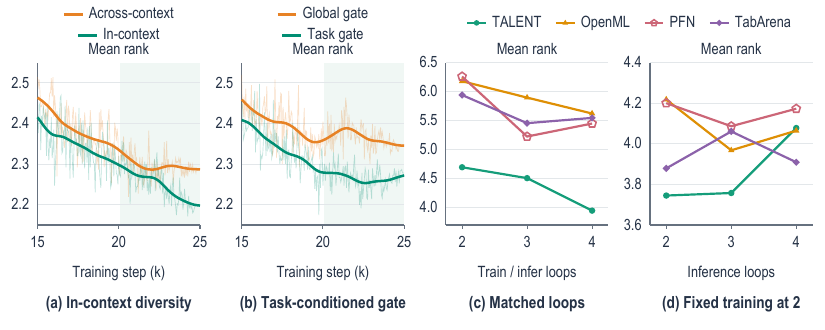}}
\vspace{-8mm}
\caption{Component analyses: (a) in-context versus cross-context diversity; (b) task-specific versus global gating; (c) matched training/inference depth; (d) extra inference loops after two-loop training. In (a,b), each variant is ranked by accuracy against four fixed baselines (TabICLv1, TabICLv2, LimiX-2M, and LimiX-16M) on each dataset, with average ranks for ties before averaging across datasets. In (c), ranks are computed on all datasets of each benchmark. Lower ranks are better; thin lines show checkpoints, thick lines show smoothed trends.}
\label{fig:component-analysis}
\label{fig:loop-depth}
\label{fig:gate-rank-trajectory}
\label{fig:rw-e4-rank-trajectory}
\vspace{-2mm}
\end{figure}

\begin{wrapfigure}{r}{0.3\textwidth}
\vspace{-5mm}
\centering
\includegraphics[width=\linewidth]{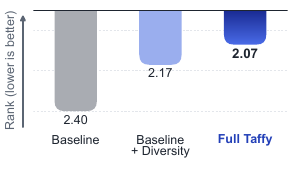}
\vspace{-8mm}\caption{Average-rank ablation on TALENT classification datasets. Each variant is ranked separately against TabICLv1, TabICLv2, LimiX-2M, and LimiX-16M.}
\label{fig:component-ablation}
\vspace{-7mm}
\end{wrapfigure}
\textbf{Ablation Study.} To verify that each proposed component contributes to \method{}, we compare three configurations in Figure~\ref{fig:component-ablation}: \emph{Baseline}, which removes both the \emph{In-Context Diversity Prior} and the \emph{Task-Conditioned Looped Transformer}; \emph{+ Diversity}, which adds only the diversity prior; and \emph{Full Taffy}, which includes both components and uses four loops during training and inference. Lower average rank indicates better performance. According to the ablation studies,
adding in-context diversity improves the rank from 2.40 to 2.17, and incorporating the task-conditioned loop further reduces it to 2.07. This monotonic improvement shows that both additions contribute to the full model, while the final comparison evaluates the loop and task gate jointly rather than isolating them.

\subsection{Analysis of Context Construction and Refinement}
\label{sec:context-analysis}
\label{sec:within-context-diversity-analysis}
\label{sec:loop-performance}
\label{sec:conditioning-analysis}
\label{sec:task-vs-global-gate}

We further analyze context organization, task-specific gating, the loop depth during training and inference, and investigate which information the task-specific gate can capture.

\textbf{In-Context Diversity vs. Cross-Context Diversity.}
To assess whether multiple environments are more useful when presented jointly within a prediction task, we compare in-context diversity with cross-context diversity. The cross-context variant retains environments generated through distribution shifts and interventions but shuffles the environment-specific data across training tasks, ensuring that each task contains data from only one environment. The in-context variant instead combines multiple related environments within half of its training tasks. Both variants use the same single-loop backbone and are evaluated every 50 steps on the TALENT classification datasets. Figure~\ref{fig:component-analysis}(a) shows the smoothed mean-rank trajectories. In-context diversity attains a lower mean rank at step 25,000, with its advantage becoming increasingly pronounced as training progresses. These results show that environmental diversity is more effective within prediction tasks than across training tasks.

\textbf{Task-Specific vs. Global Gating.}
To evaluate the contribution of task-specific statistics, we compare a task-specific gate with a global gate. Specifically, both variants are trained and evaluated with two loops on the TALENT classification datasets; the global-gate configuration retains the same model parameterization but does not receive task-specific support statistics, making $\alpha$ invariant across datasets. Figure~\ref{fig:component-analysis}(b) shows that the task-specific gate achieves a consistently lower mean rank than the global gate toward the end of training, with the gap becoming more pronounced as training progresses. This result demonstrates that support-derived task statistics provide useful signals for adapting the strength of contextual refinement across datasets. Appendix~\ref{app:gate-ablation} provides a frozen-checkpoint intervention against a support-independent coefficient and $\alpha=1$.

\textbf{Further Analysis of the Task-Conditioned Looped Transformer.}
We first vary the number of loops jointly during training and inference. Figure~\ref{fig:component-analysis}(c) shows that Taffy-4L obtains the lowest average ranks on TALENT and OpenML, whereas Taffy-3L performs best on PFN and TabArena. Increasing the matched depth from two to four loops improves the rank from 4.70 to 3.95 on TALENT and from 6.18 to 5.62 on OpenML, while PFN and TabArena favor three loops. Thus, additional trained loops can improve prediction, although the preferred depth varies across benchmark suites.

We next fix training at two loops and vary only the number of inference loops. Figure~\ref{fig:component-analysis}(d) shows that a third inference loop improves the OpenML rank from 4.22 to 3.97 and the PFN rank from 4.20 to 4.09, but does not improve TALENT or TabArena. Across the 428 benchmark entries shared by all three inference configurations, two and three loops both obtain aggregate ranks that round to 4.00, whereas four loops worsen the rank to 4.23. Additional inference-time repetition therefore does not reliably reproduce the benefit of training with greater loop depth.

\textbf{What Does the Task-Specific Gate Capture?}
To understand what the task-specific gate captures, we compute correlations between $\alpha$ and task properties, including feature composition, numerical distribution shape, label structure, and missingness, across TALENT tasks.

\begin{wraptable}[8]{r}{0.48\linewidth}
\vspace{-2.2\baselineskip}
\centering
\caption{Spearman correlations between $\alpha$ and task attributes. $^{*}$ indicates corrected $q<0.05$.}
\label{tab:gate-structure}
\begingroup
\footnotesize
\setlength{\tabcolsep}{2.6pt}
\renewcommand{\arraystretch}{1.10}
\begin{tabular*}{\linewidth}{@{\extracolsep{\fill}}l|ccc@{}}
\toprule
Property & 2L & 3L & 4L \\
\midrule
Categorical fraction & $-.533^{*}$ & $-.453^{*}$ & $-.452^{*}$ \\
Excess kurtosis & $+.230^{*}$ & $+.326^{*}$ & $+.323^{*}$ \\
Label entropy & $+.084$ & $+.010$ & $-.011$ \\
Missing ratio & $+.021$ & $+.042$ & $+.031$ \\
\bottomrule
\end{tabular*}
\vspace{-20mm}
\endgroup
\end{wraptable}

\textit{Feature composition.}
Across all three depths, a higher fraction of categorical input features is associated with lower $\alpha$ (Table~\ref{tab:gate-structure}), controlling for support rows, effective feature columns, class count, and support-label entropy. The relationship remains consistent across preprocessing views and within datasets containing both numerical and categorical features.

\textit{Numerical distributions.}
Controlling also for categorical fraction, median excess kurtosis across numerical features correlates positively with $\alpha$. Both survive multiple-testing correction, suggesting the gate responds to feature composition and numerical distribution shape.

\textit{Uncaptured task properties.}
The support-label entropy and the raw weighted missing-value ratio show no stable association with $\alpha$. Specifically, the correlations with support-label entropy remain close to zero across all three depths. A more complex conditioning architecture is needed to exploit such dataset-specific information more fully, which is an interesting future direction.

\subsection{Pretraining Data Efficiency}
\label{sec:data-efficiency}

Figure~\ref{fig:main-results}(d) compares pretraining data efficiency on TALENT classification datasets. The horizontal axis measures cumulative maximum feature elements, $\mathcal{B}(s)=\sum_{t=1}^{s}B_tR_t^{\max}d_t^{\max}$, where $B_t$ is the global task batch size, $R_t^{\max}$ is the maximum number of support and query rows per task, and $d_t^{\max}$ is the maximum number of feature columns, excluding labels. Budgets accumulate across pretraining stages without multiplying by loop count (Appendix~\ref{app:pretraining-budget}).
These results reveal two patterns. First, over the shared budget range, Taffy-3L achieves data efficiency comparable to TabICLv2 and continues to improve after TabICLv2 reaches its best result, while Taffy-4L attains higher accuracy at comparable budgets. This suggests that Taffy can extract additional predictive gains from more pretraining data after the baseline has largely saturated. Second, Taffy-4L $>$ Taffy-3L $>$ Taffy-2L holds at checkpoints at the same training steps. This consistent ordering shows that increasing the trained loop depth improves the predictive performance obtained from a fixed data budget.
\vspace{-2mm}

%% file: Taffy_conclusion.tex
\FloatBarrier
\section{Conclusion}
\vspace{-3mm}
\label{sec:conclusion}
We presented \method{} to improve how tabular foundation models learn and refine predictive relationships from context. Rather than treating task construction and context processing separately, \method{} coordinates them: the \emph{In-Context Diversity Prior} places related environments within a pretraining task to expose informative variation, while the \emph{Task-Conditioned Looped Transformer} repeatedly integrates this evidence and uses support statistics to regulate the final refinement. Controlled comparisons favor in-context diversity, task-specific gating, and greater trained loop depth. Across six classification and five regression suites, the reported classification and regression configurations attain the lowest average rank on all eleven benchmarks under the stated protocols. Overall, the results identify context composition and task-adaptive iterative refinement as complementary design axes for tabular foundation models.

%% file: Taffy_appendix.tex
\section{Synthetic Task Generation}
\label{app:environment-operators}

This section details the In-Context Diversity Prior of Section~\ref{sec:prior}. All environments of a multi-environment task are derived from one base structural causal model. Three distribution shifts change how rows are sampled from this model while keeping its graph and mechanisms fixed; two interventions change the model itself before data are generated. Table~\ref{tab:environment-operators} summarizes the five operators. Appendix~\ref{app:context-assembly} describes how the resulting environments are assembled into a single support--query task.

\begin{table}[h]
\centering
\caption{Environment operators of the In-Context Diversity Prior. The three distribution shifts reweight rows sampled from the unchanged base model, with strength fixed by $\mathrm{KL}(p_e\,\|\,p_0)=0.1$. The two interventions modify the mechanisms of $5\%$ of the non-label nodes and regenerate the data.}
\label{tab:environment-operators}
\begingroup
\footnotesize
\setlength{\tabcolsep}{3pt}
\renewcommand{\arraystretch}{1.15}
\begin{tabular}{@{}l >{\raggedright\arraybackslash}p{5.1cm} >{\raggedright\arraybackslash}p{1.7cm} >{\raggedright\arraybackslash}p{2.1cm} >{\raggedright\arraybackslash}p{1.3cm}@{}}
\toprule
Operator & Implementation & Preserved & Changed & Strength \\
\midrule
\multicolumn{5}{@{}l}{\textit{Distribution shifts (row sampling from $\mathcal{M}_0$)}} \\
Label shift & Reweight rows by $a_e(y)=\exp\!\big(\tau_e u_e(y)\big)$ & $p(x\mid y)$ & $p(y)$, hence $p(y\mid x)$ & KL $=0.1$ \\
Covariate shift & Reweight rows by $b_e(x)=\exp\!\big(-\tau_e\|x-\mu_e\|_2\big)$ & $p(y\mid x)$ & $p(x)$ & KL $=0.1$ \\
Conditional shift & Reweight rows within each class by $c_e(x,y)=\exp\!\big(-\tau_e\|x-\mu_{e,y}\|_2\big)$ & $p(y)$ & $p(x\mid y)$, hence $p(y\mid x)$ & KL $=0.1$ \\
\midrule
\multicolumn{5}{@{}l}{\textit{Interventions (mechanism changes in $\mathcal{M}_0$)}} \\
Hard intervention & Set $z_j:=v_{e,j}$ with $v_{e,j}\sim\mathcal{U}[\ell_j,u_j]$ for $j\in I_e$ & Other mechanisms & $z_j$ and its descendants & $5\%$ of non-label nodes \\
Soft intervention & Replace $g_j$ by $g_{e,j}$ from the same function family with resampled parameters, for $j\in I_e$ & Graph $G$, other mechanisms & Mechanisms in $I_e$ and their descendants & $5\%$ of non-label nodes \\
\bottomrule
\end{tabular}
\endgroup
\end{table}

\subsection{Base Causal Prototype}
\label{app:base-causal-prototype}
Let $\mathcal{M}_0$ denote a structural causal model sampled from the causal prior of \citet{hollmann2023tabpfn}, with an acyclic graph $G$ over nodes $V=\{1,\ldots,m\}$ and structural equations
\begin{equation}
    z_j=g_j\!\left(z_{\operatorname{pa}_G(j)},\epsilon_j\right),
    \qquad j=1,\ldots,m,
    \label{eq:causal-prototype}
\end{equation}
where $\epsilon=(\epsilon_1,\ldots,\epsilon_m)$ follows the sampled exogenous-noise distribution. A readout selects the feature nodes and the label node $y\in V$; nodes that are not selected remain unobserved. Evaluating Equation~\eqref{eq:causal-prototype} in topological order and applying the readout induces the base joint distribution $p_0(x,y)$, from which the base table is drawn. We use $p$ for a probability density or mass function, as appropriate. All environments of a task share $G$, the node ordering, and the readout; they differ only in their sampling distributions or in selected mechanisms.

\subsection{Distribution Shifts}
\label{app:distribution-shifts}

\paragraph{Candidate pool and weights.}
For a distribution-shift environment $e$, we draw a fresh candidate pool $\mathcal{C}_e=\{(x_i,y_i)\}_{i=1}^{n}$ from $\mathcal{M}_0$ without modifying its graph, mechanisms, noise distribution, or readout. The empirical base distribution $\hat p_0$ places mass $1/n$ on each row. For label and covariate shift, each row receives a normalized sampling weight of the exponential-tilting form
\begin{equation}
    w_{e,i}(\tau)
    =\frac{\exp\!\big(\tau\, s_e(x_i,y_i)\big)}{\sum_{k=1}^{n}\exp\!\big(\tau\, s_e(x_k,y_k)\big)},
    \qquad \tau\geq 0,
    \label{eq:tilting-weights}
\end{equation}
where the score $s_e$ depends on the shift type and is specified below. Conditional shift applies the same form separately within each class (Appendix~\ref{app:conditional-shift}). The rows of environment $e$ are then drawn with replacement from $\mathcal{C}_e$ according to $w_e(\tau_e)$, which defines the shifted distribution $p_e$. Setting $\tau=0$ gives uniform weights and recovers the base distribution.

\paragraph{Shift strength.}
We control the strength of each distribution shift through its Kullback--Leibler divergence from the base distribution. For a reweighted distribution $p_e=r_e\,p_0$ with density ratio $r_e$, $\mathrm{KL}(p_e\,\|\,p_0)=\mathbb{E}_{p_e}[\log r_e]$. On the candidate pool this becomes
\begin{equation}
    \widehat{\mathrm{KL}}_e(\tau)
    =\sum_{i=1}^{n}w_{e,i}(\tau)\log\!\big(n\,w_{e,i}(\tau)\big),
    \label{eq:empirical-kl}
\end{equation}
computed with the natural logarithm. We set a common target $\kappa=0.1$ nats for all three shift types and choose $\tau_e$ by bisection so that $\widehat{\mathrm{KL}}_e(\tau_e)=\kappa$ up to a numerical tolerance. The bisection is well posed because $\widehat{\mathrm{KL}}_e(0)=0$, $\widehat{\mathrm{KL}}_e$ is continuous, and its derivative
\begin{equation}
    \frac{\mathrm{d}}{\mathrm{d}\tau}\widehat{\mathrm{KL}}_e(\tau)
    =\tau\,\mathrm{Var}_{w_e(\tau)}\!\big[s_e(X,Y)\big]\geq 0
    \label{eq:kl-monotone}
\end{equation}
makes it nondecreasing in $\tau$. For conditional shift, the same argument applies within each class, so the class-averaged divergence is also nondecreasing. As $\tau\to\infty$, $\widehat{\mathrm{KL}}_e(\tau)$ approaches $\log(n/n_e^{\star})$, where $n_e^{\star}$ is the number of rows attaining the maximal score, so the target is attainable whenever $\log(n/n_e^{\star})>\kappa$. If the target is not attainable, we discard the environment and sample a new one.

Because each shift multiplies $p_0$ by a factor that depends only on the modified component, the joint divergence equals the divergence of that component. For label shift, $\mathrm{KL}(p_e\,\|\,p_0)=\mathrm{KL}\big(p_e(y)\,\|\,p_0(y)\big)$; for covariate shift, $\mathrm{KL}(p_e\,\|\,p_0)=\mathrm{KL}\big(p_e(x)\,\|\,p_0(x)\big)$; and for conditional shift, $\mathrm{KL}(p_e\,\|\,p_0)=\mathbb{E}_{p_0(y)}\big[\mathrm{KL}\big(p_e(x\mid y)\,\|\,p_0(x\mid y)\big)\big]$. A common $\kappa$ therefore gives the three shift types comparable magnitudes, measured on the component each one changes. The value $\kappa=0.1$ corresponds to a mild shift; for example, it moves a balanced binary label distribution from $50/50$ to approximately $72/28$.

\subsubsection{Label Shift}
\label{app:label-shift}
Label shift uses a score that depends only on the label, $s_e(x,y)=u_e(y)$, where $u_e$ assigns each class $k$ an independent score $u_{e,k}$ drawn from a uniform distribution. Because $\tau_e$ is calibrated to the KL target, the range of this uniform distribution does not affect the resulting weights. For regression, the continuous label is first discretized into 10 bins, each bin is treated as a class, and label shift and conditional shift are then constructed exactly as for classification.
The resulting label weight $a_e(y)=\exp\!\big(\tau_e u_e(y)\big)$ defines
\begin{equation}
    p_e^{\mathrm{lab}}(x,y)
    =\frac{a_e(y)p_0(x,y)}{Z_e^{\mathrm{lab}}},
    \qquad
    p_e^{\mathrm{lab}}(y)
    =\frac{a_e(y)p_0(y)}{Z_e^{\mathrm{lab}}},
    \qquad
    Z_e^{\mathrm{lab}}=\mathbb{E}_{p_0}[a_e(Y)].
    \label{eq:label-shift}
\end{equation}
Since the weight depends only on the label, $p_e^{\mathrm{lab}}(x\mid y)=p_0(x\mid y)$ wherever $p_e^{\mathrm{lab}}(y)>0$. The class proportions change, and $p(y\mid x)\propto a_e(y)\,p_0(y\mid x)$ changes accordingly. Classes with larger $u_e(y)$ are over-represented, and $\tau_e$ is calibrated so that $\mathrm{KL}\big(p_e^{\mathrm{lab}}(y)\,\|\,\hat p_0(y)\big)=0.1$.

\subsubsection{Covariate Shift}
\label{app:covariate-shift}
Covariate shift uses a score that depends only on the features. Features are standardized before distances are computed. We draw a center $\mu_e$ uniformly at random within the range of the standardized features and set $s_e(x,y)=-\|x-\mu_e\|_2$, so that $b_e(x)=\exp\!\big(-\tau_e\|x-\mu_e\|_2\big)$ and
\begin{equation}
    p_e^{\mathrm{cov}}(x,y)
    =\frac{b_e(x)p_0(x,y)}{Z_e^{\mathrm{cov}}},
    \qquad
    p_e^{\mathrm{cov}}(x)
    =\frac{b_e(x)p_0(x)}{Z_e^{\mathrm{cov}}},
    \qquad
    Z_e^{\mathrm{cov}}=\mathbb{E}_{p_0}[b_e(X)].
    \label{eq:covariate-shift}
\end{equation}
Rows closer to $\mu_e$ receive larger weights, so the environment concentrates on a random region of the feature space. The conditional label distribution satisfies $p_e^{\mathrm{cov}}(y\mid x)=p_0(y\mid x)$ on the support of the shifted feature distribution, because selection depends on the features alone.

\subsubsection{Conditional Shift}
\label{app:conditional-shift}
Conditional shift changes the feature distribution within each class while keeping the class proportions fixed \citep{zhang2015multisource}. For regression, the label bins of Appendix~\ref{app:label-shift} serve as classes, so the distribution over bins is preserved. For each class $k$, we draw a separate center $\mu_{e,k}$ in the same way as for covariate shift and set $s_e(x,y)=-\|x-\mu_{e,y}\|_2$, giving $c_e(x,y)=\exp\!\big(-\tau_e\|x-\mu_{e,y}\|_2\big)$.
The weights are normalized separately within each class,
\begin{equation}
    q_e(x\mid y)
    =\frac{c_e(x,y)p_0(x\mid y)}{Z_e^{\mathrm{cond}}(y)},
    \qquad
    Z_e^{\mathrm{cond}}(y)
    =\mathbb{E}_{p_0(X\mid y)}[c_e(X,y)],
    \label{eq:conditional-reweighting}
\end{equation}
and the environment distribution is $p_e^{\mathrm{cond}}(x,y)=p_0(y)\,q_e(x\mid y)$. On the candidate pool, row $i$ receives weight $\hat p_0(y_i)\,c_e(x_i,y_i)/\sum_{l:\,y_l=y_i}c_e(x_l,y_l)$. Within-class normalization is essential: a single global normalizer would generally change the label marginal as well. Consequently, $p_e^{\mathrm{cond}}(y)=p_0(y)$, while $p(x\mid y)$ and hence $p(y\mid x)$ change. The temperature $\tau_e$ is shared across classes and calibrated so that the class-averaged divergence $\sum_k\hat p_0(k)\,\mathrm{KL}\big(q_e(\cdot\mid k)\,\|\,\hat p_0(\cdot\mid k)\big)$ equals $0.1$. The invariances in this subsection hold for the environment distributions; finite sampled datasets need not satisfy their empirical counterparts exactly.

\subsection{Interventions}
\label{app:interventions}

\paragraph{Intervention targets.}
Interventions are applied directly to the causal graph and mechanisms of $\mathcal{M}_0$, rather than to a completed table \citep{pearl2009causality}. For each interventional environment $e$, we sample the target set $I_e$ uniformly at random without replacement from the non-label nodes $V\setminus\{y\}$, with
\begin{equation}
    |I_e|=\big\lceil 0.05\,(m-1)\big\rceil ,
    \label{eq:intervention-budget}
\end{equation}
so that at least one node is intervened on. Targets may be observed feature nodes or unobserved nodes. The label node is never intervened on directly, although the label can change through intervened ancestors. After the mechanisms in $I_e$ are modified, we draw fresh exogenous noise, evaluate the modified model in topological order, and apply the base readout. Intervention effects therefore propagate to all descendants of $I_e$, including the label when an intervened node is one of its ancestors. Unlike the distribution shifts, interventions are not calibrated to a KL target; their effect is determined by the $5\%$ budget and by the positions of the targets in $G$.

\subsubsection{Hard Interventions}
\label{app:hard-interventions}
A hard intervention replaces the structural equation of each target with a constant:
\begin{equation}
    z_j=
    \begin{cases}
        v_{e,j}, & j\in I_e,\\
        g_j(z_{\operatorname{pa}_G(j)},\epsilon_j), & j\notin I_e,
    \end{cases}
    \qquad
    v_{e,j}\sim\mathcal{U}[\ell_j,u_j],
    \label{eq:hard-intervention}
\end{equation}
where $\ell_j$ and $u_j$ are the minimum and maximum values of node $j$ in the base table. The value $v_{e,j}$ is drawn once per environment and shared by all of its rows. Each target no longer depends on its parents, so its incoming edges are removed in the intervened graph. Its outgoing effects remain: descendants are generated from their original structural equations using the intervened value. The exogenous-noise distribution and all non-target mechanisms are unchanged. When a target is a feature node, that feature is constant within the environment, which introduces a point mass into the pooled context.

\subsubsection{Soft Interventions}
\label{app:soft-interventions}
A soft intervention replaces the mechanism of each target while retaining its parental dependence:
\begin{equation}
    z_j=
    \begin{cases}
        g_{e,j}(z_{\operatorname{pa}_G(j)},\epsilon_j), & j\in I_e,\\
        g_j(z_{\operatorname{pa}_G(j)},\epsilon_j), & j\notin I_e.
    \end{cases}
    \label{eq:soft-intervention}
\end{equation}
The new mechanism $g_{e,j}$ belongs to the same function family as $g_j$ in the causal prior, with its parameters resampled independently for environment $e$. The graph $G$, the parent set of each target, and the noise distribution are unchanged, so a soft intervention changes how the parents determine the target without removing their influence \citep{brouillard2020differentiable}. All other mechanisms retain their base definitions.

\subsection{Context Assembly and Model Inputs}
\label{app:context-assembly}
For a multi-environment task, we first sample the number of environments $D$ uniformly from $\{2,3,4\}$. The environments consist of the unmodified base environment and $D-1$ further environments, each constructed with one operator sampled uniformly from the five candidates in Table~\ref{tab:environment-operators}.
Given the support and query budgets $n_s$ and $n_q$, every environment contributes an equal number of rows, $|S_e|\approx n_s/D$ and $|Q_e|\approx n_q/D$, so that $\sum_{e=1}^{D}|S_e|=n_s$ and $\sum_{e=1}^{D}|Q_e|=n_q$. These datasets are combined as in Equation~\eqref{eq:context-assembly}. Because each environment contributes to the support and query sets in the same proportion, both follow the same environment mixture; the construction introduces heterogeneity within a task rather than a shift between its support and query data. Half of the pretraining tasks use the unmodified single-environment process, and the remainder use this multi-environment construction.

The synthetic generator may use labels, including query labels, to construct a task. The prediction network receives only support features, support labels, and query features; query labels provide supervision. The task-conditioning branch computes its feature and label summaries from support data only. Environment identifiers, sampling weights, intervention targets, the causal graph, and its mechanisms are not provided to the prediction network.

The complete procedure for one multi-environment task is summarized below.
\begin{enumerate}[leftmargin=*,itemsep=1pt]
    \item Sample $\mathcal{M}_0$ from the causal prior and draw the base table.
    \item Sample $D$ uniformly from $\{2,3,4\}$ and, for each of the $D-1$ non-base environments, one operator uniformly from Table~\ref{tab:environment-operators}.
    \item For a distribution shift, draw a candidate pool from $\mathcal{M}_0$, compute scores $s_e$, find $\tau_e$ by bisection so that $\widehat{\mathrm{KL}}_e(\tau_e)=0.1$, and sample rows with replacement according to $w_e(\tau_e)$; if the target is not attainable, discard the environment and sample a new one.
    \item For an intervention, sample $I_e$ with $|I_e|=\lceil 0.05(m-1)\rceil$, modify the mechanisms in $I_e$ according to Equation~\eqref{eq:hard-intervention} or~\eqref{eq:soft-intervention}, and regenerate data in topological order.
    \item Split each environment's rows into $S_e$ and $Q_e$ with equal per-environment budgets, and concatenate them as in Equation~\eqref{eq:context-assembly}.
\end{enumerate}

\FloatBarrier

\section{Model Details}

\subsection{Support Statistics}
\label{app:support-statistics}

Table~\ref{tab:support-statistics} summarizes the 51-dimensional descriptor used by the task-specific support gate. The scalar $H$ denotes the input feature width before grouping, including padding, while $d\leq H$ counts valid features. Each of the ten per-feature summaries is aggregated across valid features by its mean, standard deviation, minimum, and maximum. Both standard deviations use the population definition. The near-zero, near-integer, and near-binary tests are $|x|<10^{-6}$, $|x-\operatorname{round}(x)|<10^{-4}$, and $\min(|x|,|x-1|)<10^{-4}$, respectively. Summaries use finite support values; a valid feature with none receives zero summaries. Padded features are excluded, and nonfinite descriptor entries are replaced by zero.

\begin{table}[!htbp]
  \centering
  \caption{Statistics used by the task-specific support gate. The descriptor contains $51$ entries. Here, $H$ is the input feature width before grouping, including padding, and $d\leq H$ is the number of valid features. Feature summaries are aggregated across valid features by mean, standard deviation, minimum, and maximum.}
  \label{tab:support-statistics}
  \vspace{3pt}
  \normalsize
  \setlength{\tabcolsep}{6pt}
  \renewcommand{\arraystretch}{1.12}
  \begin{tabular*}{\linewidth}{@{\extracolsep{\fill}}p{0.21\linewidth}p{0.64\linewidth}r@{}}
    \toprule
    \textbf{Category} & \textbf{Statistics} & \textbf{Dim.} \\
    \midrule
    \textbf{Size and validity} &
    {\raggedright $\log(1+d)$, $\log(1+n_s)$, $\log(1+N)$, $n_s/N$, $d/H$, $1-d/H$, and the fraction of finite support values over valid features.\par} & 7 \\
    \addlinespace[5pt]
    \textbf{Support labels} &
    {\raggedright Observed class count divided by $K_{\max}$; label entropy divided by $\log K_{\max}$; largest and smallest proportions among observed classes, with $K_{\max}=20$.\par} & 4 \\
    \addlinespace[5pt]
    \textbf{Support features} &
    {\raggedright Per-feature mean, standard deviation, mean absolute value, minimum, maximum, and range.\par} & $6\times4$ \\
    \addlinespace[3pt]
    & {\raggedright Per-feature fractions of zero, near-zero, near-integer, and near-binary values.\par} & $4\times4$ \\
    \midrule
    \textbf{Total} & & \textbf{51} \\
    \bottomrule
  \end{tabular*}
\end{table}

\FloatBarrier

\section{Experimental Details and Additional Results}
\label{app:experimental-details}

\subsection{Baseline Roster}
\label{app:baseline-roster}
The 20 baselines in Table~\ref{tab:main-performance} are organized into four groups. Only available results under the matching evaluation protocol are reported; inclusion in this roster does not indicate that every method has a completed result on every suite.

\paragraph{Tree-based methods.}
These comprise CatBoost \citep{prokhorenkova2018catboost}, XGBoost \citep{chen2016xgboost}, random forests (RF) \citep{breiman2001random}, and extremely randomized trees (ET) \citep{geurts2006extremely}.

\paragraph{Deep tabular methods.}
We include SwitchTab \citep{wu2024switchtab}, T2G-Former \citep{yan2023t2g}, TANGOS \citep{jeffares2023tangos}, TabCaps \citep{chen2023tabcaps}, TabM \citep{gorishniy2025tabm}, TabNet \citep{arik2021tabnet}, TabR \citep{gorishniy2024tabr}, and TabTransformer \citep{huang2020tabtransformer}. TabCaps is evaluated only for classification.

\paragraph{Tabular foundation models.}
We include TabICLv1 \citep{qu2025tabicl}, LimiX-2M \citep{wang2026limix2m}, TabPFN2 \citep{hollmann2025accurate}, Mitra \citep{zhang2025mitra}, LimiX-16M \citep{zhang2025limix}, TabICLv2 \citep{qu2026tabiclv2}, and TabPFN3 \citep{grinsztajn2026tabpfn3}. TabICLv1 is evaluated only for classification.

\paragraph{Ensemble methods.}
AutoGluon \citep{erickson2020autogluon} represents automated ensembling in the comparison roster.

\subsection{Data Splits and Preprocessing}
\label{app:experimental-preprocessing}
We retain the provided train--test splits. TALENT training and validation datasets are combined as support, BCCO uses its published splits, and OpenML-derived tasks use repeat 0 and fold 0. Numerical missing values are imputed with support-dataset medians. Categorical encodings are fitted on support examples, assigning unseen query categories a value of $-1$. Regression-label transformations are also fitted on support data, and predictions are mapped back to the original label units before evaluation. Benchmark suites may overlap.

\subsection{Training and Inference Settings}
\label{app:experimental-implementation}
\paragraph{Classification pretraining.}
Classification pretraining uses 64 AMD MI210 GPUs, a global batch size of 1,024 synthetic tasks, and a maximum context length of 4,096 rows, including support and query examples. The configured schedule comprises 25,000 steps with the Muon optimizer, a peak learning rate of $6\times10^{-4}$, weight decay of $0.01$, gradient clipping at 10, and label smoothing of $0.02$. The learning rate warms up for 500 steps and then follows cosine decay.

\paragraph{Regression model and pretraining.}
Following TabICLv2 \citep{qu2026tabiclv2}, we train a separate \method{} model for regression; it shares no parameters with the classification model. The regression model retains the column and row encoders and the task-conditioned looped Transformer of Sections~\ref{sec:encoders} and~\ref{sec:loop}, with the regression-specific components of TabICLv2. Continuous support labels are embedded with linear layers instead of class lookup tables, both in the column encoder and before the looped ICL learner. The prediction head outputs 999 quantiles at probability levels $\{0.001,0.002,\ldots,0.999\}$ instead of class probabilities, is trained with the pinball loss of Section~\ref{sec:problem-formulation}, and produces a point prediction by averaging the predicted quantiles. In the task-conditioning branch, the four class-specific support-label statistics are set to zero (Section~\ref{sec:loop}). Regression pretraining uses the In-Context Diversity Prior of Appendix~\ref{app:environment-operators} and the same hardware, global batch size, maximum context length, number of steps, and optimizer settings as classification pretraining, with approximately the same training time (Section~\ref{sec:setup}); label smoothing applies only to classification.

\paragraph{Primary-comparison model.} In Table~\ref{tab:main-performance}, classification uses Taffy-4L on all suites; regression uses the separately trained regression model. No per-dataset model selection is performed. \par\paragraph{Inference.}
In-context foundation models predict query labels from the labeled support dataset without updating pretrained parameters. The recorded cross-suite classification protocol uses 32 estimators and FP32 inference for the TabICL-based classifiers, with both identity and power-transformed feature representations. For regression, we use 8 estimators, and the ensemble construction and the feature and target preprocessing follow the TabICLv2 regression configuration: ensemble members differ in random column permutations and feature preprocessing, target transformations are fitted on the support data (Appendix~\ref{app:experimental-preprocessing}), and the predictions of the ensemble members are averaged. TabPFN and LimiX retain their model-specific inference configurations, including ensemble and retrieval settings. Thus, frozen-parameter inference does not imply an identical inference-compute budget across models.

\subsection{Pretraining Cost}
\label{app:training-cost}
Table~\ref{tab:training-cost} reports the pretraining cost of the single-pass backbone and the three \method{} variants on 64 AMD MI210 GPUs. The single-pass backbone uses the TabICLv2 architecture with one pass of the ICL stack. We measure the wall-clock time per training step and project the total time of the 25,000-step schedule from it; GPU-hours multiply this projected time by the 64 GPUs. Pretraining cost increases with loop depth: Taffy-4L requires $1.80\times$ the backbone time per step, while regression has comparable cost under the same setup (Appendix~\ref{app:experimental-implementation}).

\begin{table}[h]
\centering
\caption{Pretraining cost on 64 AMD MI210 GPUs. Time per step is measured; total time and GPU-hours are projected for the 25,000-step schedule. \emph{Relative} is the time per step divided by that of the single-pass backbone.}
\label{tab:training-cost}
\small
\begin{tabular}{@{}lrrrr@{}}
\toprule
Model & Time/step (s) & Relative & 25k steps (h) & GPU-hours \\
\midrule
Single-pass backbone & 7.01 & $1.00\times$ & 48.66 (2.03 days) & 3{,}114 \\
Taffy-2L & 8.79 & $1.25\times$ & 61.03 (2.54 days) & 3{,}906 \\
Taffy-3L & 10.62 & $1.51\times$ & 73.74 (3.07 days) & 4{,}719 \\
Taffy-4L & 12.62 & $1.80\times$ & 87.67 (3.65 days) & 5{,}611 \\
\bottomrule
\end{tabular}
\end{table}

\subsection{Metrics and Aggregation}
\label{app:experimental-metrics}
We report test accuracy for classification and RMSE for regression. Methods are ranked on each dataset, with ties assigned their average rank, and these ranks are averaged within each benchmark suite; lower average rank is better. Missing results are neither imputed nor treated as losses. Elo ratings are computed from observed dataset-level pairwise outcomes as detailed in Appendix~\ref{app:elo-comparison}.

\subsection{Primary Evaluation Pool}
\label{app:primary-evaluation-pool}
TabICLv2 \citep{qu2026tabiclv2} reports excluding development tasks but does not provide a machine-readable list. We therefore use an operational reconstruction from public materials: classification tasks identified in TabICL Appendix~F \citep{qu2025tabicl} and regression tasks matched between the TabPFNv2 supplementary development roster \citep{hollmann2025accurate} and TALENT by name, row count, and feature count. Matching dataset identities are removed from other suites as well; name matches with different source identities and table structures are retained. No missing result is interpolated.

The reconstructed classification exclusions are \texttt{allbp}, \texttt{baseball}, \texttt{delta\_ailerons}, \texttt{eye\_movements}, \texttt{eye\_movements\_bin}, \texttt{heloc}, \texttt{hill-valley}, \texttt{JapaneseVowels}, \texttt{led24}, \texttt{national-longitudinal-survey-binary}, \texttt{page-blocks}, \texttt{ringnorm}, \texttt{rl}, \texttt{thyroid-ann}, and \texttt{waveform-5000}. The regression exclusions are \texttt{analcatdata\_supreme}, \texttt{cpu\_act}, \texttt{cpu\_small}, \texttt{debutanizer}, \texttt{delta\_elevators}, \texttt{mauna-loa-atmospheric-co2}, \texttt{puma32H}, \texttt{stock\_fardamento02}, \texttt{treasury}, \texttt{weather\_izmir}, and \texttt{wind}.

After these exclusions, the primary evaluation uses the remaining memberships in each suite. Classification additionally removes TALENT tasks with more than ten classes from Table~\ref{tab:main-performance} and reports them separately below.

\subsection{Many-Class TALENT Tasks}
\label{app:many-class-evaluation}
For TALENT classification datasets with more than ten classes, we apply hierarchical classification following TabICL \citep{qu2025tabicl} together with mixed-radix ensembling \citep{qu2026tabiclv2}. To avoid imputation, Table~\ref{tab:talent-many-class} includes only the nine methods with observed results on all such datasets. Taffy has the best average rank and Elo in this matched comparison.
\input{talent_gt10_rank_elo.tex}

\subsection{LimiX-Compatible Evaluation Pool}
\label{app:limix-compatible-evaluation}
Following the dataset-eligibility criteria of LimiX-2M \citep{wang2026limix2m}, we retain datasets with at most 50,000 training examples and 10,000 features, and for classification at most 10 classes. We further remove \texttt{eucalyptus} memberships with known evaluation-sample mismatches. Table~\ref{tab:limix-compatible-rank} recomputes accuracy- and RMSE-based ranks using the complete applicable-method subset in each suite. This analysis adopts the LimiX-2M dataset filter, but not its multi-metric aggregation. Taffy has the lowest overall average rank in this pool.

\begin{table}[!t]
\centering
\caption{Average ranks on the LimiX-2M-eligible, sample-compatible evaluation pool (lower is better). Rank aggregation uses the complete applicable-method subset within each suite; dashes denote incomplete task-type coverage.}
\label{tab:limix-compatible-rank}
\begingroup
\fontsize{7.2}{8.0}\selectfont
\setlength{\tabcolsep}{1.25pt}
\renewcommand{\arraystretch}{0.90}
\resizebox{\linewidth}{!}{%
\begin{tabular}{l|*{6}{c}|*{5}{c}|c}
\toprule
\multirow{2}{*}{\textbf{Model}} & \multicolumn{6}{c|}{\textbf{Classification}} & \multicolumn{5}{c|}{\textbf{Regression}} & \multirow{2}{*}{\textbf{Average}} \\
\cmidrule(lr){2-7}\cmidrule(lr){8-12}
 & BCCO & OpenML & PFN & TALENT & TabArena & TabZilla & BCCO & CTR23 & PFN & TALENT & TabArena & \\
\midrule
\multicolumn{13}{c}{\itshape Ensemble methods} \\
\midrule
AutoGluon & 10.53 & 8.29 & 9.88 & 10.00 & 9.98 & 8.52 & 5.86 & 5.85 & 5.96 & 6.00 & 6.15 & 7.91 \\
\midrule
\multicolumn{13}{c}{\itshape Tree-based methods} \\
\midrule
CatBoost & 10.47 & 9.88 & 9.30 & 10.72 & 10.78 & 11.02 & 7.12 & 7.39 & 7.25 & 6.99 & 7.77 & 8.97 \\
XGBoost & 11.68 & 10.52 & 9.96 & 11.71 & 11.03 & 11.09 & 11.28 & 10.55 & 10.93 & 10.29 & 10.92 & 10.91 \\
ET & 12.03 & 11.56 & 10.93 & 13.23 & 14.77 & 11.80 & 10.14 & 10.03 & 10.29 & 9.70 & 9.54 & 11.27 \\
RF & 10.80 & 13.46 & 11.39 & 13.06 & 10.48 & 13.15 & 10.60 & 11.48 & 10.32 & 10.06 & 10.54 & 11.39 \\
\midrule
\multicolumn{13}{c}{\itshape Deep tabular methods} \\
\midrule
SwitchTab & 16.71 & 16.45 & 16.16 & 17.45 & 19.00 & 16.02 & 16.90 & 16.36 & 16.39 & 16.56 & 17.23 & 16.84 \\
T2G-Former & 16.49 & 17.70 & 19.00 & 16.94 & 14.95 & 17.06 & 13.18 & 13.88 & 14.43 & 13.41 & 14.23 & 15.57 \\
TANGOS & 14.26 & 16.16 & 16.93 & 15.82 & 14.39 & 14.93 & 12.10 & 11.18 & 12.43 & 12.80 & 13.00 & 14.00 \\
TabCaps & 14.78 & 14.02 & 14.46 & 14.04 & 14.84 & 13.48 & --- & --- & --- & --- & --- & 14.27 \\
TabM & 12.31 & 11.67 & 12.30 & 11.30 & 13.00 & 11.37 & 16.00 & 15.52 & 15.18 & 15.61 & 17.08 & 13.76 \\
TabNet & 18.92 & 18.34 & 19.34 & 18.70 & 18.88 & 18.93 & 16.20 & 15.58 & 16.96 & 16.41 & 16.38 & 17.69 \\
TabR & 11.99 & 11.09 & 11.14 & 11.00 & 12.12 & 10.54 & 10.66 & 10.67 & 12.93 & 11.20 & 12.31 & 11.42 \\
TabTransformer & 16.54 & 13.67 & 13.34 & 15.83 & 16.14 & 14.09 & 18.56 & 17.55 & 16.57 & 17.57 & 17.92 & 16.16 \\
\midrule
\multicolumn{13}{c}{\itshape Tabular foundation models} \\
\midrule
TabICLv1 & 7.74 & 8.16 & 7.93 & 7.63 & 7.06 & 8.65 & --- & --- & --- & --- & --- & 7.86 \\
LimiX-2M & 6.88 & 6.56 & 6.14 & 6.65 & 5.72 & 6.20 & 10.02 & 8.88 & 7.75 & 9.24 & 7.15 & 7.38 \\
TabPFN2 & 7.30 & 7.87 & 6.68 & 6.40 & 6.44 & 7.33 & 5.14 & 6.30 & 5.50 & 6.05 & 6.15 & 6.47 \\
Mitra & 9.71 & 12.60 & 12.20 & 11.34 & 10.31 & 12.09 & 9.26 & 10.39 & 8.93 & 10.13 & 9.92 & 10.63 \\
LimiX-16M & 6.09 & 6.03 & 7.00 & 5.94 & 6.03 & 6.67 & 8.16 & 7.55 & 7.57 & 7.99 & 6.46 & 6.86 \\
TabICLv2 & 5.50 & 6.20 & 5.57 & 4.70 & 5.88 & 6.41 & 3.72 & 3.73 & 3.39 & 4.12 & 3.23 & 4.77 \\
TabPFN3 & 5.83 & 6.10 & 6.45 & 4.68 & 4.73 & 6.48 & 3.12 & 4.18 & 3.93 & 3.39 & 2.69 & 4.69 \\
\textbf{\method{}} & \textbf{4.44} & \textbf{4.68} & \textbf{4.89} & \textbf{3.85} & \textbf{4.45} & \textbf{5.19} & \textbf{1.98} & \textbf{2.94} & \textbf{3.29} & \textbf{2.47} & \textbf{1.31} & \textbf{3.59} \\
\bottomrule
\end{tabular}%
}
\endgroup
\end{table}

\subsection{Pretraining Data Budget}
\label{app:pretraining-budget}
We measure the configured input-data budget by cumulative maximum feature elements, $\mathcal{B}(s)=\sum_{t=1}^{s}B_tR_t^{\max}d_t^{\max}$, where $B_t$ is the global number of training tasks per step, $R_t^{\max}$ is the maximum number of support and query rows, and $d_t^{\max}$ is the maximum number of input-feature columns, excluding labels. Multistage budgets include the initializing checkpoint's accumulated budget. Loop count does not multiply this input-data proxy, which is not a measure of computational cost or wall-clock throughput. Figure~\ref{fig:main-results}(d) compares measured checkpoints at matched or lower input-data budgets; inference settings are not fully controlled across models.

\input{elo_appendix.tex}

\subsection{Gate Analysis Details}
\label{app:gate-analysis}

\paragraph{Protocol and cohorts.}
We analyze one late-training checkpoint of each of Taffy-2L, Taffy-3L, and Taffy-4L. We reconstruct $\alpha$ on CPU from the saved FP32 gate parameters and the implementation's statistics of the full processed training-plus-validation support dataset. Input-shape metadata retain the full query count. For each dataset, we average the coefficients from the \texttt{none} and \texttt{power} preprocessing views. These are reconstructed coefficients, not recorded GPU traces. The analysis uses TALENT datasets with at most 20 classes and at most one million support cells. Categorical fraction requires matching raw metadata, while the kurtosis analysis additionally requires numerical features and finite metadata.

\paragraph{Descriptors and adjustment.}
Categorical fraction is the fraction of original input columns that are categorical. Numerical excess kurtosis is the median across raw numerical columns of $\operatorname{mean}[(x-\bar{x})^4]/\sigma^4-3$, using the population standard deviation and assigning zero to constant columns. The raw numerical summaries use finite values from a deterministic sample of at most 10{,}000 original training rows. Datasets without numerical columns are excluded from the kurtosis analysis. Raw metadata and processed support statistics therefore describe different stages of the data pipeline. We compute partial Spearman correlations by residualizing the midranks of each descriptor and $\alpha$ against an intercept and the midranks of log support rows, log effective columns, class count, and support-label entropy $H(y)/\log 20$. Kurtosis fits also control for categorical fraction.

\paragraph{Uncertainty and multiple testing.}
Table~\ref{tab:gate-adjusted-details} reports pointwise 95\% percentile intervals from 2{,}000 dataset bootstrap resamples, recomputing ranks and residuals each time. Categorical-fraction $q$-values use 9{,}999 residual permutations and Benjamini--Hochberg correction over 15 tests: three descriptors across five late-training checkpoints of the three variants. Kurtosis $q$-values use approximate two-sided $t$ tests and a separate 18-test correction covering three additional descriptors, each tested against three gate coefficients and three fixed-checkpoint inference-depth gains. Across the three model depths, Table~\ref{tab:gate-adjusted-details} shows negative associations with categorical fraction and positive associations with numerical excess kurtosis.

\begin{table}[!htbp]
\centering
\small
\setlength{\tabcolsep}{4pt}
\caption{Adjusted associations between $\alpha$ and TALENT dataset properties. Categorical fraction and numerical excess kurtosis use separate multiple-testing corrections.}
\label{tab:gate-adjusted-details}
\begin{tabular}{@{}llrr@{}}
\toprule
Descriptor & Model & Partial $\rho$ [95\% CI] & $q$ \\
\midrule
Categorical fraction & Taffy-2L & $-0.533\;[-0.652,-0.394]$ & $0.0003$ \\
 & Taffy-3L & $-0.453\;[-0.580,-0.299]$ & $0.0003$ \\
 & Taffy-4L & $-0.452\;[-0.579,-0.296]$ & $0.0003$ \\
\addlinespace
Numerical excess kurtosis & Taffy-2L & $0.230\;[0.057,0.383]$ & $0.0338$ \\
 & Taffy-3L & $0.326\;[0.167,0.466]$ & $0.000703$ \\
 & Taffy-4L & $0.323\;[0.167,0.463]$ & $0.000703$ \\
\bottomrule
\end{tabular}
\end{table}

\subsubsection{Support Rows and Effective Columns}
\label{app:gate-size-analysis}
Across TALENT datasets, $\alpha$ is positively associated with support-row count but shows no consistent association with effective column count (Table~\ref{tab:gate-size-details}).

\begin{table}[!htbp]
\centering
\small
\setlength{\tabcolsep}{4pt}
\caption{Unadjusted associations between $\alpha$ and support size on TALENT datasets. All $q$-values use Benjamini--Hochberg correction over the original 216 tests.}
\label{tab:gate-size-details}
\begin{tabular}{@{}lrrrr@{}}
\toprule
 & \multicolumn{2}{c}{Support rows} & \multicolumn{2}{c}{Effective columns} \\
\cmidrule(lr){2-3}\cmidrule(l){4-5}
Model & $\rho$ & $q$ & $\rho$ & $q$ \\
\midrule
Taffy-2L & $0.192$ & $0.0374$ & $0.138$ & $0.199$ \\
Taffy-3L & $0.297$ & $2.31\!\times\!10^{-4}$ & $0.053$ & $0.660$ \\
Taffy-4L & $0.397$ & $3.44\!\times\!10^{-7}$ & $0.022$ & $0.881$ \\
\bottomrule
\end{tabular}
\end{table}

\subsection{Frozen-Checkpoint Gate Intervention}
\label{app:gate-ablation}

We examine task-specific gating with separately trained two-loop models in Figure~\ref{fig:component-analysis}(b) and a frozen-checkpoint intervention that holds all model parameters fixed.

Starting from a Taffy-4L checkpoint, we rerun the complete inference path under three conditions: the native task-specific coefficient $\alpha_S$, the checkpoint's support-independent coefficient obtained by setting $b_S=0$, and $\alpha=1$, which bypasses the learned modulation and retains the final recurrent update at full strength. All conditions use identical parameters, four passes, preprocessing, ensemble construction, and query batches; only the update coefficient changes. Complete outputs for all three conditions are available on matched TALENT classification datasets, so we compute the comparison without imputing missing conditions. On each dataset, the three conditions are ranked by classification performance, with average ranks assigned to ties, before averaging across datasets. \begin{table}[!htbp] \centering \small \setlength{\tabcolsep}{7pt} \caption{Frozen-checkpoint gate intervention on matched TALENT classification datasets. Mean rank among the three inference conditions; lower is better.} \label{tab:gate-intervention-rank} \begin{tabular}{@{}lcc@{}} \toprule Inference condition & Update coefficient & Average rank $\downarrow$ \\ \midrule Task-specific gate & $\alpha_S$ & \textbf{1.57} \\ Support-independent gate & $\tanh(a)$ & 1.88 \\ Gate bypass (full update) & $1$ & 2.55 \\ \bottomrule \end{tabular} \end{table}

On the matched TALENT datasets, the task-specific gate has the lowest average rank among the three inference conditions (Table~\ref{tab:gate-intervention-rank}).  \section{Limitations} \label{app:limitations}  Task conditioning in \method{} uses a single task-level scalar gate applied to the final refinement. Although this keeps the mechanism lightweight, it cannot assign different refinement strengths across examples, features, layers, or earlier passes. Comparing this design with finer-grained gating under matched parameter and compute budgets is an important extension.

%% file: talent_gt10_rank_elo.tex
\begin{table}[t]
\centering
\caption{Results on TALENT classification datasets with more than ten classes. Only methods with observed results on all such datasets are included.}
\label{tab:talent-many-class}
\small
\begin{tabular}{lcc}
\toprule
Method & Avg. rank $\downarrow$ & Elo $\uparrow$ \\
\midrule
RF & 8.42 & 1000.00 \\
TabICLv1 & 5.17 & 1595.23 \\
LimiX-2M & 5.50 & 1552.44 \\
TabPFN2 & 5.17 & 1595.23 \\
Mitra & 8.25 & 1044.05 \\
LimiX-16M & 5.12 & 1600.49 \\
TabICLv2 & 2.62 & 1923.15 \\
TabPFN3 & 2.62 & 1923.15 \\
Taffy & \textbf{2.12} & \textbf{2000.81} \\
\bottomrule
\end{tabular}
\end{table}

%% file: elo_appendix.tex
\clearpage
\subsection{Elo Comparison}
\label{app:elo-comparison}

We report Elo ratings under both the primary evaluation pool and the LimiX-2M eligibility rule. The former follows the development-set and many-class controls used for Table~\ref{tab:main-performance}; the latter uses the stricter LimiX-2M dataset filter. Taffy has the highest mean Elo under both evaluation pools.

\paragraph{Rating protocol.}
Each dataset supplies one comparison per method pair. Higher accuracy wins for classification and lower RMSE wins for regression; ties count as half a win. Accuracy differences within $10^{-12}$ are ties; RMSE ties require equality of stored values. We jointly fit an unpenalized Bradley--Terry model per benchmark, with equal dataset weights. The fitted strengths $\theta_i$ give order-independent Elo ratings,
\begin{equation}
E_i=1000+\frac{400}{\ln 10}(\theta_i-\theta_{\mathrm{RF}}),
\qquad
p_{ij}=\frac{1}{1+10^{(E_j-E_i)/400}}.
\label{eq:elo-rating}
\end{equation}
RF is anchored at 1000, following TabArena's calibration convention \citep{erickson2025tabarena}. Average is the mean of all eleven unrounded ratings, not a pooled Elo fit; it is omitted for methods without regression results. The evaluation settings are those used for the reported results; this is not a compute-matched comparison or a test of statistical significance.

\paragraph{Primary evaluation.}
Table~\ref{tab:main-elo-dev-filtered-combined} applies the same operational development-set exclusions as Table~\ref{tab:main-performance}; classification additionally excludes TALENT tasks with more than ten classes.
\input{main_elo_combined.tex}

\paragraph{LimiX-compatible evaluation.}
Table~\ref{tab:limix-compatible-elo} uses the same LimiX-compatible evaluation pool as Table~\ref{tab:limix-compatible-rank}. Missing scores are neither imputed nor treated as losses.

\begin{table}[!t]
\centering
\caption{Elo ratings on the LimiX-2M-eligible, sample-compatible evaluation pool (higher is better; RF = 1000 per benchmark). Only observed pairwise outcomes are used; missing scores are not imputed, and dashes denote incomplete task-type coverage.}
\label{tab:limix-compatible-elo}
\begingroup
\fontsize{7.2}{8.0}\selectfont
\setlength{\tabcolsep}{1.25pt}
\renewcommand{\arraystretch}{0.90}
\resizebox{\linewidth}{!}{%
\begin{tabular}{l|*{6}{c}|*{5}{c}|c}
\toprule
\multirow{2}{*}{\textbf{Model}} & \multicolumn{6}{c|}{\textbf{Classification}} & \multicolumn{5}{c|}{\textbf{Regression}} & \multirow{2}{*}{\textbf{Average}} \\
\cmidrule(lr){2-7}\cmidrule(lr){8-12}
 & BCCO & OpenML & PFN & TALENT & TabArena & TabZilla & BCCO & CTR23 & PFN & TALENT & TabArena & \\
\midrule
\multicolumn{13}{c}{\itshape Ensemble methods} \\
\midrule
AutoGluon & 1011.38 & 1217.39 & 1064.14 & 1138.25 & 1021.54 & 1187.80 & 1303.99 & 1310.03 & 1257.72 & 1236.25 & 1332.51 & 1189.18 \\
\midrule
\multicolumn{13}{c}{\itshape Tree-based methods} \\
\midrule
CatBoost & 1013.58 & 1150.20 & 1088.56 & 1106.17 & 973.79 & 1086.68 & 1216.33 & 1219.76 & 1178.34 & 1174.86 & 1204.78 & 1128.46 \\
XGBoost & 962.97 & 1123.70 & 1061.09 & 1061.68 & 976.48 & 1084.46 & 957.85 & 1050.70 & 964.64 & 986.98 & 970.83 & 1018.31 \\
ET & 948.06 & 1080.91 & 1017.65 & 991.99 & 809.15 & 1053.96 & 1028.09 & 1078.09 & 1002.07 & 1020.32 & 1074.19 & 1009.50 \\
RF & 1000.00 & 1000.00 & 1000.00 & 1000.00 & 1000.00 & 1000.00 & 1000.00 & 1000.00 & 1000.00 & 1000.00 & 1000.00 & 1000.00 \\
\midrule
\multicolumn{13}{c}{\itshape Deep tabular methods} \\
\midrule
SwitchTab & 724.60 & 854.24 & 767.09 & 768.92 & 521.75 & 869.73 & 480.01 & 677.30 & 590.18 & 556.05 & 322.86 & 648.43 \\
T2G-Former & 737.05 & 779.84 & 555.29 & 799.91 & 800.02 & 812.06 & 831.16 & 860.26 & 743.84 & 800.85 & 682.11 & 763.85 \\
TANGOS & 849.72 & 870.70 & 719.58 & 863.75 & 827.08 & 921.98 & 905.28 & 1016.50 & 874.66 & 840.19 & 800.31 & 862.70 \\
TabCaps & 825.54 & 974.46 & 859.48 & 954.11 & 792.33 & 985.93 & --- & --- & --- & --- & --- & 898.64 \\
TabM & 935.27 & 1075.44 & 959.47 & 1079.86 & 880.95 & 1072.59 & 586.18 & 746.94 & 689.38 & 640.91 & 344.10 & 819.19 \\
TabNet & 570.05 & 734.24 & 519.59 & 675.82 & 528.89 & 680.37 & 564.30 & 742.28 & 535.86 & 569.60 & 434.53 & 595.96 \\
TabR & 949.47 & 1098.94 & 1009.99 & 1093.77 & 922.90 & 1105.94 & 996.32 & 1044.23 & 843.41 & 935.41 & 860.32 & 987.34 \\
TabTransformer & 734.37 & 990.23 & 912.61 & 863.60 & 723.74 & 959.55 & 182.08 & 555.86 & 573.90 & 444.82 & 218.03 & 650.80 \\
\midrule
\multicolumn{13}{c}{\itshape Tabular foundation models} \\
\midrule
TabICLv1 & 1131.13 & 1223.07 & 1148.49 & 1249.15 & 1152.41 & 1182.42 & --- & --- & --- & --- & --- & 1181.11 \\
LimiX-2M & 1170.90 & 1295.61 & 1232.14 & 1298.69 & 1198.36 & 1290.29 & 1035.39 & 1139.16 & 1148.70 & 1045.70 & 1251.78 & 1191.52 \\
TabPFN2 & 1151.09 & 1235.63 & 1206.00 & 1311.45 & 1182.70 & 1238.43 & 1358.90 & 1282.47 & 1288.05 & 1233.01 & 1332.51 & 1256.39 \\
Mitra & 1045.63 & 1036.81 & 964.20 & 1078.11 & 1007.40 & 1043.47 & 1081.58 & 1058.78 & 1080.25 & 996.04 & 1045.86 & 1039.83 \\
LimiX-16M & 1210.05 & 1321.67 & 1190.80 & 1336.53 & 1203.14 & 1268.48 & 1149.40 & 1211.35 & 1159.22 & 1116.35 & 1306.92 & 1224.90 \\
TabICLv2 & 1240.08 & 1313.12 & 1261.44 & 1409.58 & 1211.18 & 1280.59 & 1483.45 & 1458.84 & 1447.78 & 1370.48 & 1647.59 & 1374.92 \\
TabPFN3 & 1222.74 & 1317.99 & 1216.32 & 1410.83 & 1273.94 & 1277.10 & 1546.03 & 1423.03 & 1402.27 & 1433.48 & 1731.77 & 1386.86 \\
\textbf{\method{}} & \textbf{1301.29} & \textbf{1394.88} & \textbf{1298.76} & \textbf{1467.72} & \textbf{1290.81} & \textbf{1341.96} & \textbf{1697.50} & \textbf{1529.66} & \textbf{1457.50} & \textbf{1530.52} & \textbf{2047.85} & \textbf{1487.13} \\
\bottomrule
\end{tabular}%
}
\endgroup
\end{table}

%% file: main_elo_combined.tex
\begin{table*}[t]
\centering
\caption{Corresponding observed-only Elo ratings under the operational development-set reconstruction (higher is better).}
\label{tab:main-elo-dev-filtered-combined}
\scriptsize
\setlength{\tabcolsep}{2.6pt}
\resizebox{\textwidth}{!}{%
\begin{tabular}{l|cccccc|ccccc|c}
\toprule
& \multicolumn{6}{c|}{Classification} & \multicolumn{5}{c|}{Regression} & \\
Method & TALENT & BCCO & OpenML & PFN & TabArena & TabZilla & TALENT & BCCO & CTR23 & PFN & TabArena & Mean \\
\midrule
AutoGluon & 1139.13 & 1001.45 & 1217.39 & 1064.14 & 1022.49 & 1187.80 & 1241.40 & 1295.80 & 1310.03 & 1257.72 & 1332.51 & 1188.17 \\
CatBoost & 1105.99 & 1015.31 & 1150.20 & 1088.56 & 986.67 & 1086.68 & 1171.15 & 1210.85 & 1219.76 & 1178.34 & 1204.78 & 1128.94 \\
XGBoost & 1056.39 & 957.92 & 1123.70 & 1061.09 & 975.45 & 1084.46 & 981.73 & 956.97 & 1050.70 & 964.64 & 970.83 & 1016.72 \\
ET & 986.81 & 934.06 & 1080.91 & 1017.65 & 800.16 & 1053.96 & 1016.33 & 1027.80 & 1078.09 & 1002.07 & 1074.19 & 1006.55 \\
RF & 1000.00 & 1000.00 & 1000.00 & 1000.00 & 1000.00 & 1000.00 & 1000.00 & 1000.00 & 1000.00 & 1000.00 & 1000.00 & 1000.00 \\
SwitchTab & 769.70 & 718.50 & 854.24 & 767.09 & 522.28 & 869.73 & 573.18 & 490.49 & 677.30 & 590.18 & 322.86 & 650.50 \\
T2G-Former & 806.69 & 735.87 & 779.84 & 555.29 & 790.55 & 812.06 & 803.46 & 827.86 & 860.26 & 743.84 & 682.11 & 763.44 \\
TANGOS & 866.91 & 844.88 & 870.70 & 719.58 & 819.03 & 921.98 & 846.05 & 909.75 & 1016.50 & 874.66 & 800.31 & 862.76 \\
TabCaps & 960.88 & 822.62 & 974.46 & 859.48 & 795.37 & 985.93 & -- & -- & -- & -- & -- & 899.79 \\
TabM & 1067.57 & 935.32 & 1075.44 & 959.47 & 885.79 & 1072.59 & 652.53 & 576.99 & 746.94 & 689.38 & 344.10 & 818.74 \\
TabNet & 683.62 & 567.24 & 734.24 & 519.59 & 532.55 & 680.37 & 583.67 & 572.26 & 742.28 & 535.86 & 434.53 & 598.75 \\
TabR & 1087.05 & 947.49 & 1098.94 & 1009.99 & 926.06 & 1105.94 & 939.75 & 992.01 & 1044.23 & 843.41 & 860.32 & 986.84 \\
TabTransformer & 873.54 & 725.87 & 990.23 & 912.61 & 726.10 & 959.55 & 467.88 & 201.40 & 555.86 & 573.90 & 218.03 & 655.00 \\
TabICLv1 & 1250.62 & 1125.39 & 1223.07 & 1148.49 & 1159.11 & 1182.42 & -- & -- & -- & -- & -- & 1181.52 \\
LimiX-2M & 1287.21 & 1157.56 & 1295.61 & 1232.14 & 1229.12 & 1290.29 & 1048.98 & 1051.50 & 1139.16 & 1148.70 & 1251.78 & 1193.82 \\
TabPFN2 & 1310.96 & 1142.44 & 1235.63 & 1206.00 & 1190.78 & 1238.43 & 1223.98 & 1361.42 & 1282.47 & 1288.05 & 1332.51 & 1255.70 \\
Mitra & 1077.65 & 1039.40 & 1036.81 & 964.20 & 1007.72 & 1043.47 & 983.63 & 1079.13 & 1058.78 & 1080.25 & 1045.86 & 1037.90 \\
LimiX-16M & 1329.06 & 1199.91 & 1321.67 & 1190.80 & 1212.16 & 1268.48 & 1114.45 & 1153.87 & 1211.35 & 1159.22 & 1306.92 & 1224.35 \\
TabICLv2 & 1416.24 & 1232.42 & 1313.12 & 1261.44 & 1220.58 & 1280.59 & 1365.45 & 1480.73 & 1458.84 & 1447.78 & 1647.59 & 1374.98 \\
TabPFN3 & 1414.45 & 1216.29 & 1317.99 & 1216.32 & 1286.42 & 1277.10 & 1422.65 & 1543.54 & 1423.03 & 1402.27 & 1731.77 & 1386.53 \\
Taffy & \textbf{1476.85} & \textbf{1292.81} & \textbf{1394.88} & \textbf{1298.76} & \textbf{1304.18} & \textbf{1341.96} & \textbf{1529.52} & \textbf{1678.27} & \textbf{1529.66} & \textbf{1457.50} & \textbf{2047.85} & 1486.57 \\
\bottomrule
\end{tabular}%
}
\end{table*}